\pdfoutput=1
\documentclass{article}

\PassOptionsToPackage{numbers}{natbib}
\usepackage[preprint]{arxiv}

\usepackage[utf8]{inputenc} 
\usepackage[T1]{fontenc}    
\usepackage{hyperref}       
\usepackage{url}            
\usepackage{booktabs}       
\usepackage{amsfonts}       
\usepackage{nicefrac}       
\usepackage{microtype}      
\usepackage{comment}        
\usepackage{amsmath}
\usepackage{multirow}
\usepackage{bm}
\usepackage{amsthm}
\usepackage{hhline}
\usepackage{amssymb}
\usepackage{makecell}
\usepackage{mathtools}
\usepackage{color}
\usepackage{xcolor}
\usepackage{MnSymbol}
\usepackage{makecell}
\usepackage{graphicx}
\usepackage{arydshln}
\usepackage{booktabs}
\usepackage{framed}
\usepackage[font=small]{caption}
\usepackage[scientific-notation=true]{siunitx}
\usepackage{wrapfig}
\usepackage{lipsum}
\usepackage{sidecap}
\usepackage{pifont}
\usepackage{fixltx2e}
\usepackage[flushleft]{threeparttable}
\usepackage{diagbox}

\newcommand{\name}{\textsc{Brainish}}

\DeclareMathOperator*{\argmax}{arg\,max}

\definecolor{gg}{RGB}{15,150,15}
\definecolor{rr}{RGB}{230,45,45}

\newcommand{\PreserveBackslash}[1]{\let\temp=\\#1\let\\=\temp}
\newcolumntype{C}[1]{>{\PreserveBackslash\centering}p{#1}}
\newcolumntype{R}[1]{>{\PreserveBackslash\raggedleft}p{#1}}
\newcolumntype{L}[1]{>{\PreserveBackslash\raggedright}p{#1}}

\makeatletter
\def\maketag@@@#1{\hbox{\m@th\normalfont\normalsize#1}}
\makeatother

\usepackage{algorithm}
\usepackage[noend]{algpseudocode}
\algdef{SE}[SUBALG]{Indent}{EndIndent}{}{\algorithmicend\ }%
\algtext*{Indent}
\algtext*{EndIndent}

\AtBeginDocument{%
  \addtolength\abovedisplayskip{-0.3\baselineskip}%
  \addtolength\belowdisplayskip{-0.3\baselineskip}%
}

\title{\sc \name: Formalizing A Multimodal Language for\\Intelligence and Consciousness}

\author{%
    Paul Pu Liang\\
    Machine Learning Department\\
    Carnegie Mellon University\\
    Pittsburgh, PA 15213\\
    \href{mailto:pliang@cs.cmu.edu}{\texttt{pliang@cs.cmu.edu}}
}

\begin{document}

\maketitle

\begin{abstract}
Having a rich multimodal inner language is an important component of human intelligence that enables several necessary core cognitive functions such as multimodal prediction, translation, and generation. Building upon the Conscious Turing Machine (CTM), a machine model for consciousness proposed by~\citet{blum2021theoretical}, we describe the desiderata of a multimodal language called \name, comprising words, images, audio, and sensations combined in representations that the CTM's processors use to communicate with each other.
We define the syntax and semantics of \name\ before operationalizing this language through the lens of multimodal artificial intelligence, a vibrant research area studying the computational tools necessary for processing and relating information from heterogeneous signals. Our general framework for learning \name\ involves designing (1) unimodal encoders to segment and represent unimodal data, (2) a coordinated representation space that relates and composes unimodal features to derive holistic meaning across multimodal inputs, and (3) decoders to map multimodal representations into predictions (for fusion) or raw data (for translation or generation).
Through discussing how \name\ is crucial for communication and coordination in order to achieve consciousness in the CTM, and by implementing a simple version of \name\ and evaluating its capability of demonstrating intelligence on multimodal prediction and retrieval tasks on several real-world image, text, and audio datasets, we argue that such an inner language will be important for advances in machine models of intelligence and consciousness.
\end{abstract}

\vspace{-2mm}
\section{Introduction}
\vspace{-2mm}

Our perception of the natural world surrounding us involves multiple sensory modalities: we see objects, hear audio signals, feel textures, smell fragrances, and taste flavors. A \textit{modality} refers to a way in which a signal exists or is experienced. Multiple modalities then refer to a combination of multiple signals each expressed in heterogeneous manners~\citep{baltruvsaitis2018multimodal}. The ability to seamlessly integrate and translate between different modalities is a hallmark of human intelligence that enables core cognitive functions such as multimodal prediction, translation, and generation~\citep{kosslyn2010multimodal,meltzoff1999origins,murray2011neural,nanay2018multimodal,ohshiro2011normalization,stein2008multisensory,talsma2010multifaceted}:
\begin{enumerate}
    \item \textit{Multimodal fusion:} encoding modalities both in individuality (e.g., reading a book) as well as in context with other modalities (e.g., listening to movie dialog while watching acted facial expressions).
    \item \textit{Multimodal translation:} converting a unit from one modality to semantically corresponding units in another modality. For example, seeing an image and describing its contents in text.
    \item \textit{Multimodal generation:} parallel generation of realistic data from multiple modalities. For example, dreaming constitutes synchronized imaginations of speech, sight, touch, smell, and other modalities.
\end{enumerate}
The ability to perform multimodal processing requires the development of a \textit{multimodal language} comprising words, images, and sensations combined in representations that are understood by the brain~\cite{blum2021theoretical,kosslyn2010multimodal,lohse2021subcortical,murray2011neural,nanay2018multimodal} and decodable to human-perceptible data forms~\cite{calvert2001crossmodal,meaidi2014sensory,pearson2019human}. In this paper, we describe the desiderata of such a multimodal language called \name\ in accomplishing similar functionality in AI. We develop the underlying key principles of this multimodal language by defining its syntax (grammar) and semantics (meaning). Starting from a local level (e.g., individual words, image regions, audio segments) multimodal \textit{semantics} study the relationships across data units with common meaning expressed across multiple modalities. Multimodal \textit{syntax} then defines the compositional structure that jointly builds up shared multimodal units to derive holistic meaning at a global level (e.g., an entire video). Together, a multimodal language comprising syntax and semantics enables us to effectively (1) fuse modalities by discovering complementary information unique to each signal, (2) translate between modalities by taking advantage of common meaning across signals, and (3) generate new multimodal data starting with co-occurring local units and composing them to form global data of rich content.

We next describe how to operationalize our formalism of \name\ through the lens of multimodal machine learning. Multimodal machine learning has emerged as a vibrant research area studying the computational tools necessary for processing and relating information from heterogeneous signals. Building upon recent work, our general framework for capturing unimodal and multimodal syntax and semantics in \name\ involves designing (1) suitable \textit{unimodal encoders} to segment and represent unimodal data, (2) a \textit{coordinated representation space} that relates and composes unimodal features to derive holistic meaning across entire multimodal inputs, and (3) \textit{decoders} to map multimodal representations into either a prediction (for fusion) or raw data (for translation or generation).

We evaluate this proposed framework in $2$ ways: first conceptually by discussing how the \name\ multimodal language is crucial for communication and coordination in the Conscious Turing Machine (CTM), a machine model for \textit{consciousness} as proposed by~\citet{blum2021theoretical}, and then by implementing a simple version of \name\ and evaluating its capability of demonstrating \textit{intelligence} on a suite of multimodal prediction and retrieval tasks on real-world image~\cite{krizhevsky2009learning}, text~\cite{yummly_dataset}, and audio~\cite{piczak2015esc} datasets. We conclude by arguing that a multimodal language is central to the study of intelligence and consciousness in human and artificial intelligence.

For neuroscientists, we hope that this paper can introduce several challenges and opportunities from the perspective of multimodal machine learning which can inspire computational models of AI based on human intelligence~\citep{blum2021theoretical,chella2007artificial,mcdermott2007artificial,moscovitch1995models,seth2007models}. For computer scientists, we hope that the insights from human intelligence and consciousness can potentially inform the design of new computational datasets, algorithms, and evaluation frameworks~\cite{bengio2017consciousness,lake2017building}.

In the following section, we first provide necessary background in multimodal machine learning (Section~\ref{sec:background}) to motivate our definition of a multimodal language (Section~\ref{sec:properties}). We then discuss algorithms for operationally learning this multimodal language (Section~\ref{sec:language}). Using these tools, we apply them to the CTM~\cite{blum2021theoretical} in Section~\ref{sec:ctm} and to a case study on real-world multimodal datasets in Section~\ref{sec:ml}.

\vspace{-2mm}
\section{Background: Multimodal Machine Learning}
\label{sec:background}
\vspace{-2mm}

We define a modality as a single particular mode in which a signal is expressed or experienced. Multiple modalities then refer to a combination of multiple heterogeneous signals~\citep{baltruvsaitis2018multimodal}. Each modality can be represented as \textit{static} inputs without a time dimension (such as images or a table of numerical data) or as \textit{temporal} inputs which come in a sequence with a time-dimension such as language (a sequence of tokens), video (a sequence of frames/audio features/optical flow features), or time-series data. Many real-world research problems are inherently multimodal: from the early research on audio-visual speech recognition~\citep{dupont2000audio} to the recent explosion of interest in language, vision, and video understanding~\citep{dupont2000audio} for applications such as multimedia~\citep{liang2018multimodal,1667983,ngiam2011multimodal}, affective computing~\citep{liang2019tensor,PORIA201798}, robotics~\citep{kirchner2019embedded,lee2019making}, finance~\citep{doi:10.1177/0170840618765019}, dialogue~\citep{Pittermann2010}, human-computer interaction~\citep{dumas2009multimodal,obrenovic2004modeling}, education~\citep{moro2019multimodal} and healthcare~\citep{medical,xu2019multimodal}. The research field of multimodal machine learning (ML) brings unique challenges for both computational and theoretical research, and has emerged as a vibrant interdisciplinary field of immense importance and with extraordinary potential~\citep{baltruvsaitis2018multimodal}. As relevant background, we review some of the core research challenges and main application areas of this research field.

\vspace{-1mm}
\subsection{Core research challenges}
\vspace{-1mm}

\begin{figure}[tbp]
\centering
\includegraphics[width=\linewidth]{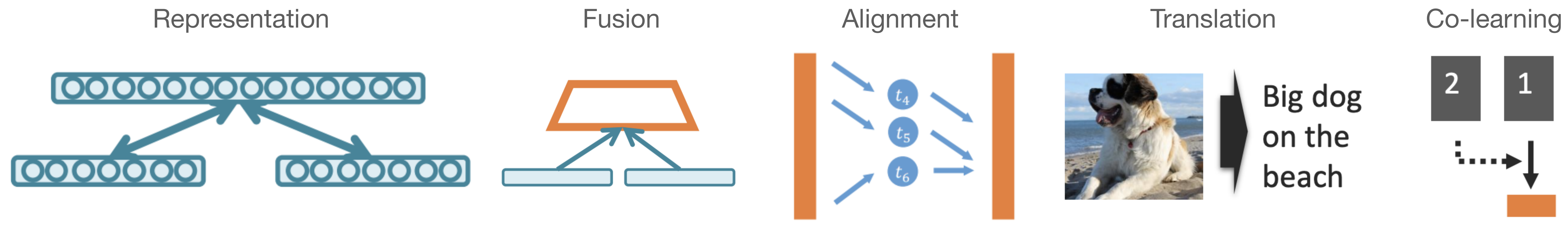}
\caption{Core research challenges in multimodal learning: \textbf{Representation} studies how to represent and summarize the multimodal data to highlight the complementarity and synchrony between modalities. \textbf{Fusion} aims to combine information from two or more modalities to perform a prediction (e.g., classification, regression). \textbf{Alignment} aims to identify the direct relations between units from two or more different modalities. \textbf{Translation} studies the generation of semantically-aligned information in a new and different modality. \textbf{Co-learning} aims to transfer knowledge between modalities and their representations. Note that these technical challenges are not mutually exclusive - solving each real-world multimodal problem typically requires tackling more than one core challenge in conjunction.}
\label{fig:overview}
\vspace{-2mm}
\end{figure}

There are several core challenges in multimodal learning. We briefly summarize a few below and give some definitions and examples, but defer the reader to a survey paper for more details~\citep{baltruvsaitis2018multimodal}. Please refer to Figure~\ref{fig:overview} for an overview of these technical challenges. In Table~\ref{table:multimodal}, we give examples of concrete machine learning tasks and datasets in each area. Note that these technical challenges are not mutually exclusive. Solving each real-world multimodal problem typically requires tackling more than one core challenge in conjunction.

\textbf{Representation:} Firstly, the challenge of multimodal \textit{representation} aims to represent and summarize the multimodal data to highlight the complementarity and synchrony between modalities. The heterogeneity of multimodal data makes it particularly challenging to learn coordinated and joint representations. For example, language is often seen as symbolic while audio and visual modalities are represented as signals. Multimodal representation learning is typically exemplified by \textit{joint representations} (integrating information from 2 or more modalities, effectively reducing the number of separate representations) and \textit{coordinated representations} (interchanging cross-modal information with the goal of keeping the same number of representations but improving multimodal contextualization). Representation is a particularly overarching challenge that needs to be considered for every more specific challenge below.

\textbf{Fusion:} In \textit{multimodal fusion}, the main challenge is to combine information from two or more modalities to perform a prediction (e.g., classification, regression). Classic examples for multimodal fusion include audio-visual speech recognition where visual lip motion is fused with speech signals to predict spoken words~\citep{dupont2000audio}, or recognizing human emotion from language, spoken speech, and visual gestures.

\textbf{Alignment:} The challenge of multimodal \textit{alignment} aims to identify the direct relations between units from two or more different modalities. For example, when analyzing the speech and gestures of a human subject, how can we align specific gestures with spoken words or utterances? Alignment between modalities is challenging since it may depend on long-range dependencies, involves ambiguous segmentation (e.g., words or utterances), and could be either one-to-one, many-to-many, or not exist at all. Some core tasks in multimodal alignment are cross-modal retrieval~\citep{wang2016comprehensive} and visual grounding~\citep{agrawal2017vqa}.

\textbf{Translation:} Multimodal fusion and alignment can be contrasted with \textit{multimodal translation} where the goal is to generate semantically-aligned information in a new and different modality~\citep{vinyals2016show}. For example, generating a descriptive caption of an image can help to improve the accessibility of visual content for blind people~\citep{gurari2018vizwiz}. Multimodal translation brings about new difficulties involving the generation of high-dimensional structured multimodal data as well as their evaluation.

\textbf{Co-learning:} Finally, a fifth challenge, \textit{co-learning}, is to transfer knowledge between modalities and their representations. Exemplified by algorithms of co-training, conceptual grounding, and zero-shot learning, how can knowledge learned from one modality (e.g., predicted labels or representation) help a computational model trained on a different modality? This challenge is particularly relevant when one of the modalities has limited resources. Some examples of co-learning involve transferring knowledge from knowledge graphs to visual classification~\citep{marino2016more}, images to machine translation~\citep{specia2016shared}, and video to language~\cite{zadeh2020foundations}.

\vspace{-1mm}
\subsection{Core Applications and Datasets}
\vspace{-1mm}

\begin{table*}[]
\fontsize{9}{11}\selectfont
\setlength\tabcolsep{3.0pt}
\vspace{-0mm}
\caption{Some representative machine learning tasks and datasets for each of the multimodal challenges of fusion, alignment, translation, and co-learning. Representation is a more overarching challenge that needs to be considered for the other more specific challenges, so it does not have specific tasks or datasets. Input modalities span $a$: audio, $e$: embodied environment, $f$: force sensor, $g$: graph, $i$: image $\ell$: language, $o$: optical flow, $p$: proprioception sensor, $s$: set, $t$: time-series, $ta$: tabular, $v$: video.}
\centering
\footnotesize
\vspace{-0mm}
\begin{tabular}{l|cccccccc}
\hline \hline
\multicolumn{1}{l|}{Area} & Task & Dataset & Modalities \\
\hline
\multirow{14}{*}{Fusion} & sarcasm prediction & \textsc{MUStARD}~\cite{castro2019towards} & $\{\ell,v,a\} \rightarrow y$ \\
& sentiment prediction & \textsc{CMU-MOSI}~\cite{zadeh2016mosi} & $\{\ell,v,a\} \rightarrow y$ \\
& humor prediction & \textsc{UR-FUNNY}~\cite{hasan2019ur} & $\{\ell,v,a\} \rightarrow y$ \\
& emotion prediction & \textsc{CMU-MOSEI}~\cite{zadeh2018multimodal} & $\{\ell,v,a\} \rightarrow y$ \\
& mortality, disease code prediction & \textsc{MIMIC}~\cite{MIMIC} & $\{t,ta\} \rightarrow y$ \\
& object pose prediction & \textsc{MuJoCo Push}~\citep{lee2020multimodal} & $\{i,f,p\} \rightarrow y$ \\
& contact, robot pose prediction & \textsc{Vision\&Touch}~\citep{lee2019making} & $\{i,f,p\} \rightarrow y$ \\
& movie genre classification & \textsc{MM-IMDb}~\citep{arevalo2017gated} & $\{\ell,i\} \rightarrow y$ \\
& digit classification & \textsc{AV-MNIST}~\citep{vielzeuf2018centralnet} & $\{i,a\} \rightarrow y$ \\
& human action classification & \textsc{Kinetics400}~\citep{kay2017kinetics} & $\{v,a,o\} \rightarrow y$ \\
& video classification & \textsc{YouTube-8M}~\citep{abu2016youtube} & $\{\ell,v,a\} \rightarrow y$ \\
& image question answering & \textsc{VQA}~\citep{agrawal2017vqa} & $\{\ell,i\} \rightarrow y$ \\
& video question answering & \textsc{TVQA}~\citep{lei2018tvqa} & $\{\ell,v\} \rightarrow y$ \\
& environment question answering & \textsc{EQA}~\citep{das2018embodied} & $\{\ell,e\} \rightarrow y$ \\
\hline \hline
\multirow{3}{*}{Alignment} & image-caption retrieval & \textsc{Flickr-30k}~\citep{plummer2015flickr30k} & $\ell \leftrightarrow i$ \\
& audio-caption retrieval & \textsc{AudioCaps}~\citep{kim2019audiocaps} & $\ell \leftrightarrow a$ \\
& audio-visual retrieval & \textsc{YouTube-8M}~\citep{abu2016youtube} & $a \leftrightarrow v$ \\
\hline \hline
\multirow{5}{*}{Translation} & image captioning & \textsc{MSCOCO}~\citep{lin2014microsoft} & $i \rightarrow \ell$ \\
& video captioning & \textsc{LSMDC}~\citep{rohrbach2017movie} & $v \rightarrow \ell$ \\
& speech recognition & \textsc{WSJ}~\citep{paul1992design} & $a \rightarrow \ell$ \\
& text-to-speech & \textsc{LibriTTS}~\cite{zen2019libritts} & $\ell \rightarrow a$ \\
& image generation & \textsc{Conceptual Captions}~\citep{ramesh2021zero,sharma2018conceptual} & $\ell \rightarrow i$ \\
\hline \hline
\multirow{3}{*}{Co-learning} & video $\rightarrow$ text & \textsc{CMU-MOSI} $\rightarrow$ \textsc{SST}~\citep{zadeh2020foundations} & $\{\ell,v,a\} \rightarrow \ell$ \\
& text $\rightarrow$ image & \textsc{GloVe} $\rightarrow$ \textsc{CIFAR10}~\citep{socher2013zero} & $\{i,\ell\} \rightarrow i$ \\
& knowledge graph $\rightarrow$ image & Visual Genome~\citep{krishna2017visual,marino2016more} & $\{i,g\} \rightarrow i$ \\
\hline \hline
\end{tabular}
\vspace{-4mm}
\label{table:multimodal}
\end{table*}

In this subsection, we list some major applications of multimodal machine learning in the real world.

\textbf{Affective computing} studies the perception of human affective states (emotions, sentiment, and personalities) from our natural display of multimodal signals spanning language (spoken words), visual (facial expressions, gestures), and acoustic (prosody, speech tone)~\citep{picard2000affective}. Some commonly studied datasets and tasks involving fusing \textit{language}, \textit{video}, and \textit{audio} time-series data to predict sentiment (\textsc{CMU-MOSI}~\citep{zadeh2016mosi} and \textsc{CMU-MOSEI}~\citep{zadeh2018multimodal}), emotions (\textsc{CMU-MOSEI}~\citep{zadeh2018multimodal}), humor (\textsc{UR-FUNNY}~\cite{hasan2019ur}), and sarcasm (\textsc{MUStARD}~\citep{castro2019towards}).

\textbf{Healthcare:} Medical decision-making often involves integrating complementary signals from several sources such as lab tests, imaging reports, and patient-doctor conversations. Multimodal models can help doctors make sense of high-dimensional data and assist them in the diagnosis process~\cite{amisha2019overview}. \textsc{MIMIC} is a large-scale dataset~\citep{MIMIC} which records ICU patient data including \textit{time-series} data measured every hour and other \textit{tabular numerical} data about the patient (e.g., age, gender, ethnicity) to predict mortality rate and the disease ICD-$9$-code.

\textbf{Robotics:} Modern robot systems are equipped with multiple sensors to aid in their decision-making. Recent work has explored methods to integrate visual (RGB and depth), force, and proprioception sensors to predict the pose of the object being pushed by the robot end-effector~\citep{lee2019making} or action-conditional learning objectives that capture forward dynamics of the different modalities (contact prediction and robot end-effector pose)~\citep{lee2019making}. These multi-sensor robots have been successfully applied into haptic robots~\citep{pai2005multisensory,seminara2019active} and surgical robots~\citep{abiri2019multi,bethea2004application}. More generally, language~\citep{luketina2019survey} and audio~\citep{dean2020see} have also emerged as useful signals in learning policies for reinforcement learning in both simulation and the real world.

\textbf{Multimedia:} A significant body of research in multimodal learning has been fueled by the large availability of multimedia data (language, image, video, and audio) on the internet. The research field of multimedia involves understanding and synthesizing different content forms into a single interactive medium. Several real-world challenges include audio-visual video classification (classifying a video into a particular genre~\citep{zhang2001audio} and recommending similar videos), image/video question answering (asking and answering text-based questions given a relevant image or video~\citep{agrawal2017vqa,lei2018tvqa}), image/video captioning (generating descriptive text for a given image or video~\citep{drossos2020clotho,vinyals2016show}), image/audio/text retrieval~\citep{mitrovic2010features,rui1999image,zhen2019deep} (retrieving relevant image, audio, video, or text articles given a search query in another modality).

\vspace{-1mm}
\subsection{Case Studies}
\vspace{-1mm}

To motivate these multimodal tasks and challenges, we illustrate examples of state-of-the-art models tackling these technical challenges through $3$ case studies:
\begin{enumerate}
    \item \textit{Video-based affect recognition} aims to predict human sentiment and emotions from spoken text, prosody, and visual gestures~\citep{zadeh2018multimodal}. This is primarily a fusion problem to combine multimodal signals to make a prediction. At the same time, a model also needs to learn suitable representations of each signal before fusion can be performed. These representations should be able to relate signals that represent similar meanings. For example, loud voices and laughter reinforce each other to predict stronger happiness over each individually. Local fusion of the loud voice and laugh signals can only be performed with the discovery of the underlying complementary information across the audio and image modalities.
    \item \textit{Image-based question answering} aims to correctly answer a text-based question in reference to a relevant image (e.g., asking \textit{what color is the table} in reference to an image depicting a brown table). This is both a fusion and alignment problem: fusion because the goal is to integrate complementary information from the image and text question, and alignment because one has to relate words in the question (e.g., \textit{table}) to a specific part of the image referencing that word.
    \item \textit{Image-caption retrieval} aims to retrieve a semantically relevant image given a text caption or search query~\citep{plummer2015flickr30k}. Similarly, in image-caption generation, the goal is to generate a caption, one word at a time, describing an image. These are both primarily translation problems with the goal of learning relationships between images and text to enable translating from one modality to another. It also requires learning alignment between image and text where units from images are close together with their semantically corresponding units in the caption.
\end{enumerate}

\vspace{-2mm}
\section{Towards Formalizing A Multimodal Language}
\label{sec:properties}
\vspace{-2mm}

In this section, we identify the underlying key principles towards formalizing a general multimodal language. We begin with a basic problem setup that defines a universe of concepts and their manifestations as multimodal data through a generative process. Using this setup, we first define the notions of unimodal syntax and semantics, before extending these definitions to capture multimodal syntax and semantics.

\textbf{Setup:} Suppose there are 2 modalities (e.g., image and text) and a set $M$ of underlying atomic abstract concepts (e.g., \textit{cats, dogs, tables, chairs}). Each modality is comprised of a set of atomic units - the most basic unit of real-world data in that modality which cannot (or rather, the user chooses to not) be broken down into further units. For example, when working with the text modality, a user may choose the level of words as the most basic unit, in which case the set of atomic units $M_1$ would be a word-level vocabulary. When working with the image modality, one might choose the level of cropped object patches as the most basic unit, which results in a `visual vocabulary' $M_2$ (e.g., cropped images of cats, dogs, tables, and chairs).

\textbf{A generative process for multimodal data:} These abstract concepts are manifested as real-world data in terms of these $2$ modalities. This manifestation process can be seen as stochastic functions mapping units from $M$ to those in $M_1$ and $M_2$. Continuing with the above example, the concept \textit{cat} could be mapped to the text modality as words \textit{cat}, \textit{feline}, \textit{kitten}, and so on. Similarly, it could be mapped to the image modality as different basic images of cats with varying colors, sizes, and features. While this may seem straightforward for object-based concepts, the generative process becomes more ambiguous when dealing with non-objects.
For example, $M$ could also contain abstract emotions such as happiness, which can be expressed in language via positive words, audio via loud voices and positive tones, and visual via smiles, laughs, eye movements, and many more. Typically, $M_1 \neq M_2 >> M$ - there are many more real-world manifestations of abstract concepts through raw data than the abstract concepts themselves. For example, there are many words describing a cat and also many possible visual scenes of a cat.

Further building on this setup, problems of significance in the real world are typically not defined directly in terms of atomic units, but rather their compositions into ordered collections. For example, instead of words and object regions, multimodal tasks involve sentences, long paragraphs, dense images, and videos~\cite{lin2014microsoft,plummer2015flickr30k,zadeh2018multimodal} as an ordered sequence of atomic units. There is typically a set of rules governing this ordered composition, such as grammar in language~\citep{chomsky2014aspects} or visual relationships in image~\citep{horn1998visual}.

\textbf{Challenges:} The core research challenges of representation, fusion, alignment, translation, and co-learning are then defined on top of multimodal data and an associated task.
These underlying concepts and compositions are important since they usually define the task space. For example, representation and fusion generally require recovering the underlying abstract concept (e.g., \textit{cats}, \textit{dogs}, \textit{tables}, \textit{chairs}, \textit{happiness}, \textit{sadness}, \textit{sarcasm}) after their manifestation into atomic concepts and composition into real-world high-dimensional multimodal data. Alignment, translation, and co-learning require the discovery of pairings across data related by shared underlying abstract concepts to enable retrieval, generation, and information transfer.
Therefore, all of these challenges require studying the relationships between data and abstract concepts (i.e., semantics), as well as the composition of atomic units into higher-order sequences (i.e., syntax). We will proceed to formalize these notions of unimodal syntax and semantics, as well as multimodal syntax and semantics in the next section. Together, they create a multimodal language necessary for modeling the generative process of multimodal data to solve associated tasks.

\vspace{-1mm}
\subsection{Unimodal Syntax and Semantics}
\vspace{-1mm}

\begin{figure}[tbp]
\centering
\vspace{-2mm}
\includegraphics[width=\linewidth]{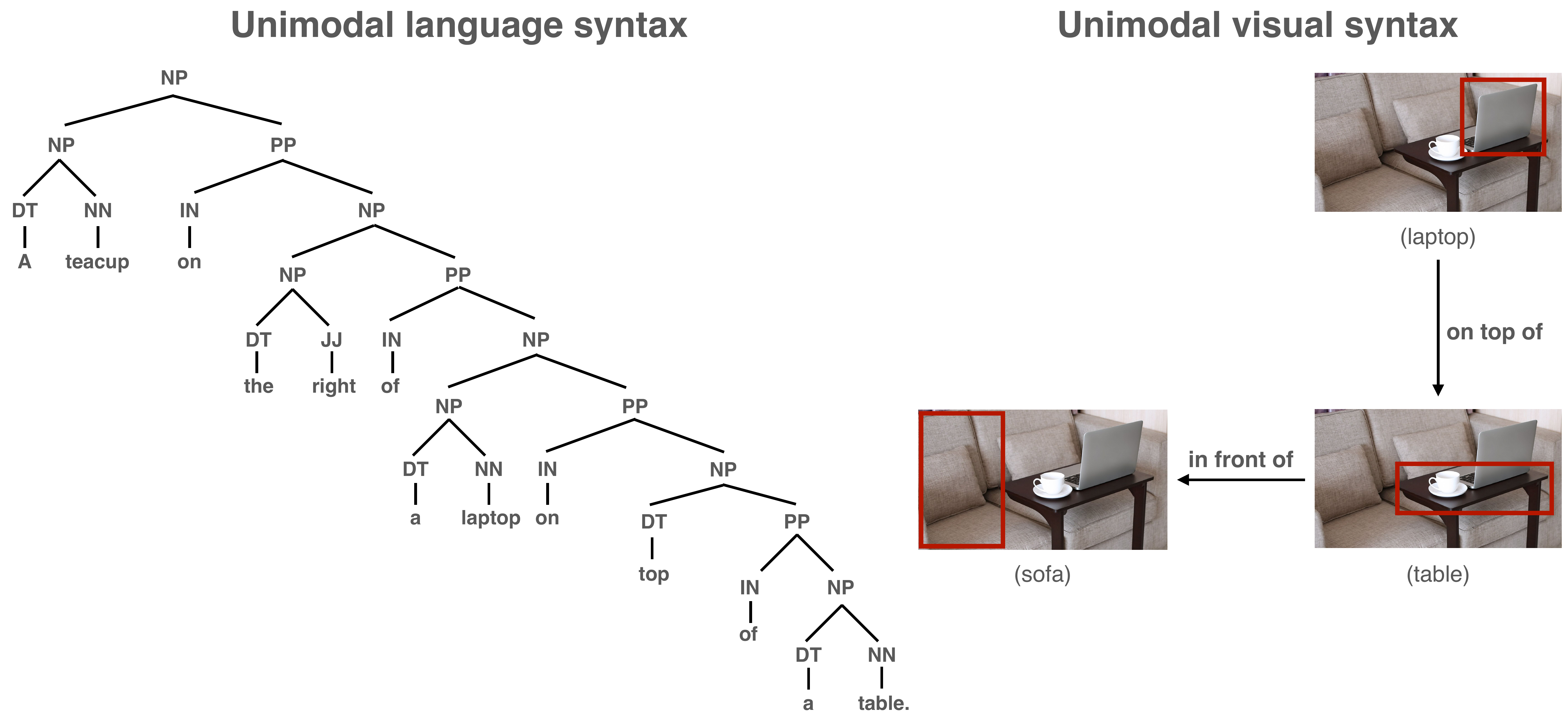}
\caption{\textbf{Unimodal syntax} of an arbitrary modality refers to the compositional structure of atomic units in that modality into more complex yet structurally valid data. Left: In language, the syntax is typically defined via a set of fixed production rules that govern how words (the atomic unit) build up into a grammatical sentence, which can then be visualized as a syntax tree. Right: In the visual modality, syntax refers to certain rules that govern the composition of individual object units into a visual scene - the laptop typically goes on top of a table rather than the sofa, and the table is typically in front of the sofa.}
\label{fig:syntax}
\vspace{-4mm}
\end{figure}

We begin with a treatment of \textbf{unimodal syntax}. Commonly studied in language, syntax refers to grammar - the set of fixed composition rules that govern how words (the atomic unit) build up into a structurally valid (i.e., grammatical) sentence~\citep{chomsky2014aspects}. The set of composition rules resulting in a grammatical sentence can then be visualized as a constituency-based parse tree (see the left side of Figure~\ref{fig:syntax}), where certain parts of speech (NP: noun phrase, VP: verb phrase, etc.) are composed according to grammar rules~\citep{carnie2021syntax}. In the visual modality, atomic units could refer to individual objects in a scene, such as a laptop, a teacup, a table, and a sofa. Visual syntax (right side of Figure~\ref{fig:syntax}) then refers to rules that govern the composition of individual object units into a visual scene~\citep{horn1998visual} - the laptop and teacup typically go onto a table rather than the sofa, and the sofa is typically parallel but lower than the table. Visual syntax is informed and constrained by spatial dimensions and perceptual principles, but there are typically no fixed rules. Instead, probabilistic rules are learned from a natural distribution of images or based on visual design principles such as gestalt theory, visual topologies, prior associations, or visual context~\citep{horn1998visual}.

More generally, the syntax of an arbitrary modality refers to the compositional structure of atomic units in that modality into more complex yet structurally valid data. Formally, given 2 subsets of atomic units $A, B \subseteq M_1$ (or $M_2$), unimodal syntax defines a composition function $f : A \times B \rightarrow [0,1]$ that outputs a value representing the validity of a particular composition. In the case of language which has a deterministic syntax, the output is either $0$ or $1$: output $0$ denotes invalid composition, output $1$ denotes valid composition. For the visual modality, a probabilistic syntax means that the output is a range $[0,1]$ representing the likelihood of a valid composition based on the natural distribution of images.

\begin{figure}[tbp]
\centering
\vspace{-4mm}
\includegraphics[width=\linewidth]{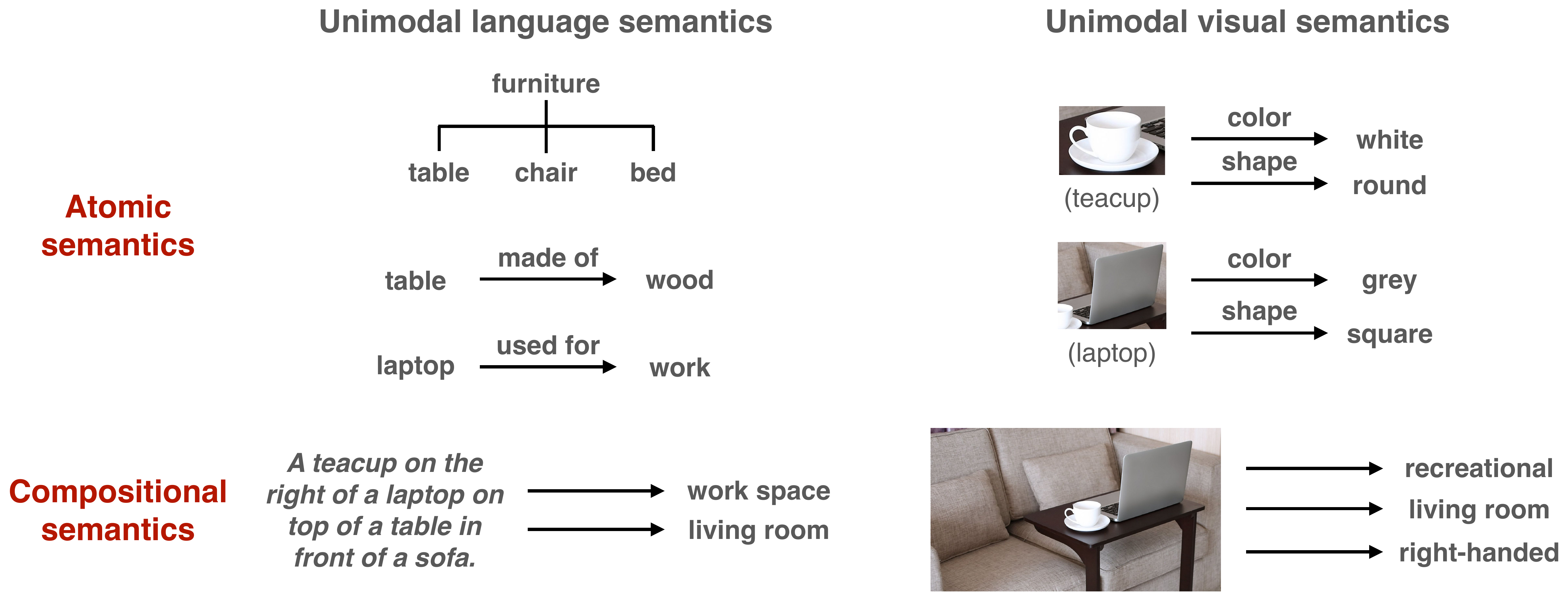}
\caption{The \textbf{unimodal semantics} of an arbitrary modality refer to the meaning of each atomic unit in that modality (atomic semantics), as well as the meaning of compositions of those units as governed by a corresponding syntax (compositional semantics). Left: The lexical semantics of each word (table, laptop) represent word meaning as exemplified through semantic hierarchies, their real-world usages, or interactions with other related objects. When composed in a sentence, one can infer higher-level concepts such as a person's work desk and whether the scene belongs to a house of an office. Right: The visual semantics of each object provides complementary information through visual properties (size, shape, and color) that might not be present in the language. Similarly, compositional semantics represent meaning from the entire visual scene.}
\label{fig:semantics}
\vspace{-4mm}
\end{figure}

The \textbf{unimodal semantics} of an arbitrary modality refers to the meaning of each atomic unit in that modality (atomic semantics), as well as the meaning of compositions of those units as governed by a corresponding syntax (compositional semantics).
In the former, unimodal atomic semantics aim to discover the meaning of each atomic unit in $M_1$ (and in $M_2$). What is meaning? In linguistics, the study of word meaning includes the study of words both locally and globally.
At local levels (i.e., only a word), meaning is communicated through the relationships between the distinct senses of a word and how words are derived~\citep{taylor2017lexical} through word-level semantic relations such as synonyms, antonyms, hypernyms, hyponyms, homonyms, and polysems~\citep{loos2004glossary}.
At global levels, meaning is communicated via how words are used in grammatical contexts~\citep{brown2005encyclopedia}.
In the visual modality, these local relationships are captured through visual properties. Following the same example above and illustrated in Figure~\ref{fig:semantics}, the visual semantics of each object (laptop, a teacup, a table, and a sofa) would typically represent both local meaning: what object it is, their physical properties (size, shape, and color), as well as global meaning: what they are used for and how they would interact with other related objects~\citep{giunchiglia2021towards}. 

In the latter, unimodal compositional semantics study how the meaning of atomic units correlates with the structure of the language or syntax (also known as syntax-semantics interface~\citep{pustejovsky1998generative}). For example, when individual object regions are composed together in a scene based on visual syntax, visual compositional semantics would then represent higher-level concepts such as a person's work desk, whether the person is right or left-handed depending on the relative position of the teacup, and whether the scene belongs to a house of an office, and so on (see right side of Figure~\ref{fig:semantics}).

In language, semantics are typically learned via the distributional hypothesis: the idea that units (i.e., words) or their compositions (i.e., sentences) of similar meaning tend to occur in the same context~\citep{harris1954distributional}. Extensions of these ideas to visual semantics have also explored how visual scenes are classified and organized into a semantic hierarchy based on the occurrence of objects in their visual context~\citep{cai2020towards,cohn2016multimodal,giunchiglia2021towards}.

\vspace{-1mm}
\subsection{Multimodal Syntax and Semantics}
\vspace{-1mm}

\begin{figure}[tbp]
\centering
\vspace{-6mm}
\includegraphics[width=\linewidth]{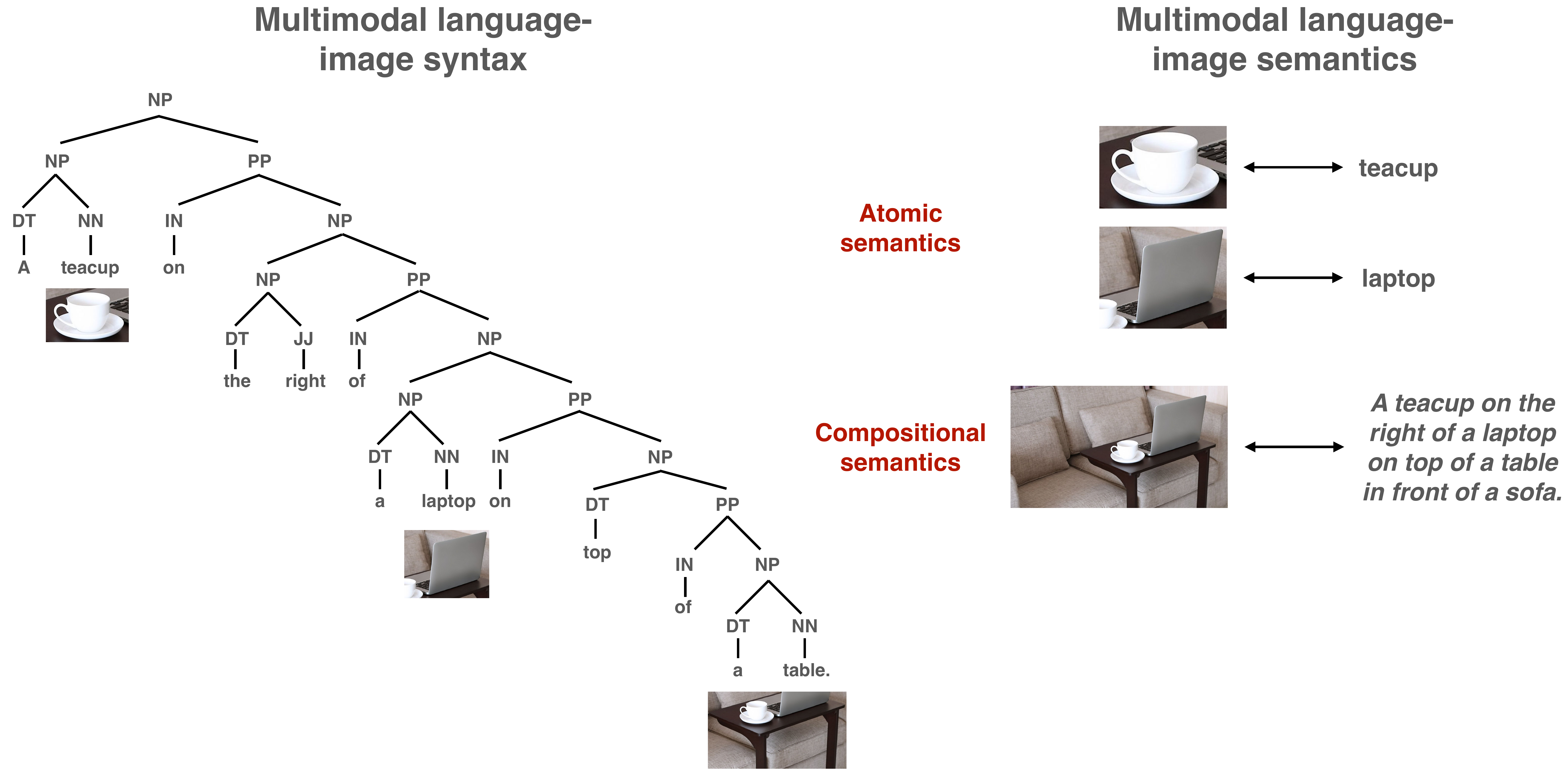}
\caption{\textbf{Multimodal syntax} refers to a compositional structure that jointly builds up multimodal units. For example, the textual description of a visual scene defines a multimodal syntax that describes exactly how one would compose individual visual objects (tea cup, laptop, table, sofa) into the complete visual scene, rather than a probabilistic visual syntax as described in Figure~\ref{fig:syntax}. \textbf{Multimodal semantics} refer to shared meaning across modalities both at atomic and compositional levels.}
\label{fig:multimodal}
\vspace{-4mm}
\end{figure}

While unimodal syntax and semantics are typically studied in unimodal machine learning (e.g., classification and generation of image or text individually), multimodal machine learning requires extending the notion of syntax and semantics to multimodal tasks, which we illustrate in Figure~\ref{fig:multimodal}. In this section, we provide preliminary definitions and examples of multimodal syntax and semantics.

\textbf{Multimodal atomic semantics} study shared meaning of atomic units across multiple modalities~\cite{kiela2014learning,kiela2015multi}. These relationships are present due to the underlying pairing across atomic units in $M_1$ and $M_2$ through abstract concepts in $M$ that generated them. Therefore, multimodal atomic semantics can be seen as extending the idea of semantic relations from within the same modality to across different modalities. For example, the semantics of a visual image of a dog should correspond to the semantics of an audio clip of a dog barking. Given $2$ atomic units $a \in M_1, b \in M_2$, learning multimodal atomic semantics, therefore, involves learning a pairing/alignment function $f : a \times b \rightarrow [0,1]$ where the output represents a likelihood of the $2$ units across both modalities having shared meaning. While there are cases where meaning is exactly shared across units, there are also cases where the matching is many-to-many (many possible dog barks for the same image of a dog), or does not exist at all (it might be hard or impossible to describe a bark exactly in words). In certain cases, the underlying abstract concepts can be expressed in ambiguous or idiosyncratic manners - the abstract concept of sarcasm is commonly expressed through positive words in the language modality yet disappointed/exasperated tones or gestures in the audio or visual modalities~\citep{altrabsheh2015detecting}. One has to identify the relationships between these atomic units of seemingly contradictory meaning in order to accurately predict sarcasm from multimodal data. Similarly, textual references to visual objects in complex scenes can possibly be ambiguous and require careful reasoning~\citep{johnson2017clevr,winograd1971procedures}.

\textbf{Multimodal syntax} involves learning the compositional structure that jointly builds up shared multimodal atomic units to derive holistic meaning. Depending on the specific problem, the compositional structure, or multimodal grammar, can fall in several cases:
\begin{enumerate}
    \item Deterministic syntax (e.g., grammar in text) helping to resolve probabilistic syntax (e.g., image). For example, on the right side of Figure~\ref{fig:multimodal}, the textual description of a visual scene \textit{``a teacup on the right of a laptop on top of a table in front of a sofa''} defines a multimodal syntax that describes exactly how one would compose individual visual objects (\textit{tea cup, laptop, table, sofa}) into the complete visual scene, rather than a probabilistic visual syntax as described in Figure~\ref{fig:syntax} (probabilistic since the spatial relationships between \textit{tea cup, laptop, table, sofa} are not exactly determined). In this case, additional deterministic information from the language modality helps resolve ambiguity in the composition of visual objects into a scene.

    \item Joint temporal syntax. In cases where deterministic rules are not present (e.g., image and audio modalities), the joint compositional structure has to be learned from naturally occurring data. One type of multimodal syntax is a common shared temporal dimension across modalities. This is exemplified in video data, where a common time dimension builds up units at individual time steps. For example, after learning local atomic pairs (smile, loud voice) and (closed eyes, laughter), it is likely that the smile happens at the same time as closed eyes which co-occur as a result of laughter, all of which happens before a loud voice. Jointly composing paired units can help reduce ambiguity by decreasing the space of all possible configurations.
\end{enumerate}
At a global level, \textbf{multimodal compositional semantics} study how meaning is built up under the compositional structure defined by multimodal syntax. Similar to unimodal compositional semantics, we show examples of compositional meaning across multimodal data on the right side of Figure~\ref{fig:multimodal}: a complete visual scene matching a complete description of the scene in language. Composing individual local relationships results in a global representation of multimodal data. Each modality often provides additional information for a task which could come in the following forms~\citep{baltruvsaitis2018multimodal}:
\begin{enumerate}
    \item Joint information is present in both modalities that reinforce each other (e.g., loud voice and smile). In this case, existing information is contextualized and reinforced based on other modalities.
    \item Complementary information is present in one modality but not the other (e.g., monotone voice, but positive words). Existing information is new and necessary since it is not present in other modalities.
\end{enumerate}
Together, a multimodal language comprising syntax and semantics enables us to effectively (1) \textit{fuse} modalities by discovering complementary information unique to each signal, (2) \textit{translate} between modalities by taking advantage of joint information across signals, and (3) \textit{generate} new multimodal data starting with co-occurring local units and composing them to form global data of rich content.

\vspace{-2mm}
\section{Operationally Learning a Multimodal Language}
\label{sec:language}
\vspace{-2mm}

Based on the previous treatment of syntax and semantics in unimodal and multimodal tasks, our goal is to operationally learn a multimodal language in practice. This section details our framework for multimodal language learning and references to current research on the machine learning side.

\vspace{-1mm}
\subsection{A Framework for Multimodal Language Learning}
\vspace{-1mm}

\begin{figure}[tbp]
\centering
\includegraphics[width=\linewidth]{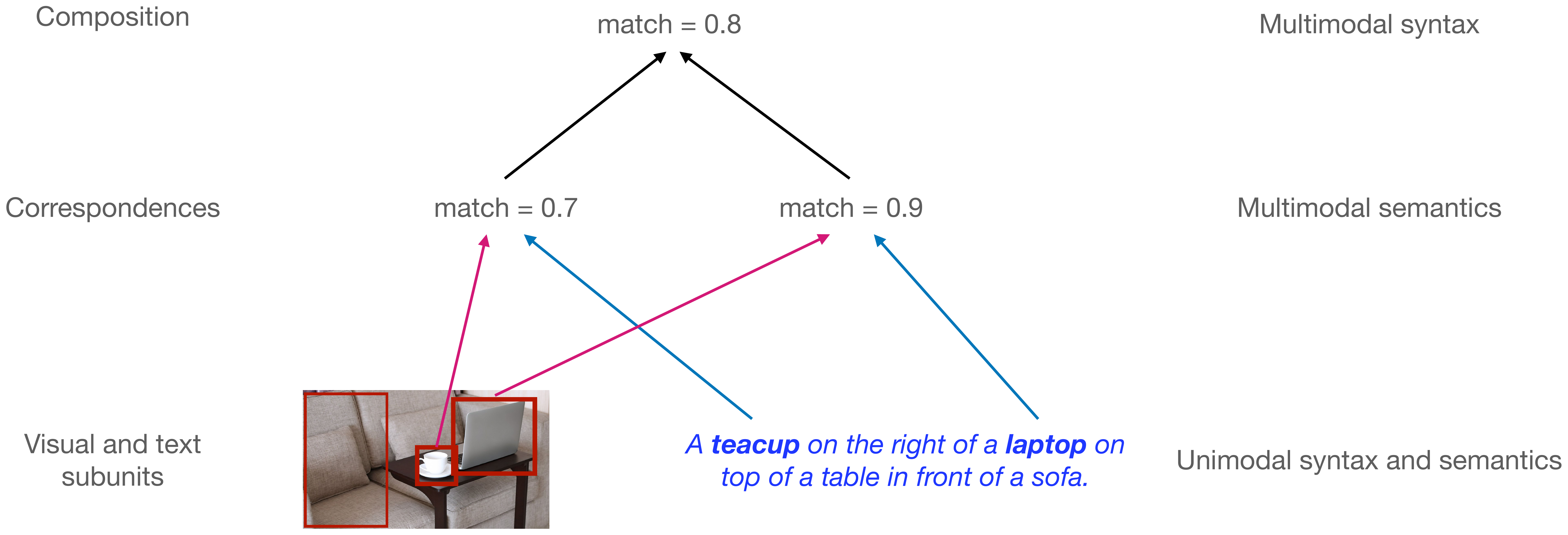}
\caption{We propose a \textbf{general multimodal language} for processing multimodal data to discover unimodal and multimodal syntax and semantics essential to a range of challenges in multimodal machine learning. This generalization involves discovering atomic units in each modality, their correspondences within and across modalities, and their composition for a specific prediction task.}
\label{fig:method}
\vspace{-4mm}
\end{figure}

Our proposed framework consists of $3$ steps: designing encoders from multimodal data to representations, learning a suitable representation space, and designing decoders from representations back into data. Each of these steps is designed to capture unimodal and multimodal syntax and semantics.
\begin{enumerate}
    \item \textit{Encoders} take in raw data from different modalities and model unimodal syntax and semantics into a unimodal representation. Syntax is captured by segmenting each modality into atomic units, and semantics are captured by learning a representation summarizing the meaning of each atomic unit. By modeling unimodal syntax and semantics, the result is a fine-grained unimodal feature representation capturing both compositionality and meaning in unimodal data.
    
    \item The \textit{representation space} takes in multiple unimodal feature representations across modalities and captures multimodal syntax and semantics. Multimodal semantics are learned via alignment: the matching between atomic units across multiple modalities based on shared meaning. Multimodal syntax involves learning how aligned subsets of atomic units compose to derive holistic meaning across entire multimodal inputs (rather than at the level of units). The result is a coordinated multimodal representation capturing shared and composed meaning across multimodal inputs.
    
    \item Finally, \textit{decoders} take in multimodal representations and output a prediction, which can either be a classification label in prediction tasks or raw data in generation tasks. In the former, fused multimodal data is important to capture complementary information for prediction (e.g., predicting emotion from language, speech, and gestures). In the latter, generation can be in the same modality (e.g., dialog prediction in language) or different modality (describing an image in language), all of which necessitate starting from a coordinated
    multimodal representation.

\end{enumerate}
To motivate this learning process, we show how they would be executed in three case studies, and show an example in Figure~\ref{fig:method}:
\begin{enumerate}
    \item \textit{Video-based affect recognition} aims to predict human sentiment and emotions from spoken text, prosody, and visual gestures~\citep{zadeh2018multimodal}. Unimodal encoders segment and learn atomic units such as units of speech (e.g., a specific word or phrase), tone (e.g., loud voice or speaking quickly), and gestures (e.g., a yawn or laugh). Learning correspondences between these units then refers to estimating a fused representation based on unimodal, bimodal, and trimodal interactions (e.g., loud voice and laugh reinforce each other to predict stronger happiness over each individually). Finally, the decoder composes these individual local predictions across the entire video to form a global video-level emotion prediction.
    \item \textit{Image-caption retrieval} aims to retrieve a semantically relevant image given a text caption or search query~\citep{plummer2015flickr30k}. Atomic units could refer to specific objects or certain nouns or phrases in text, in which case learning correspondences between these units refers to estimating an alignment probability based on semantic similarity. Finally, these individual object and phrase-level alignments are composed to form a global image-text alignment estimate.
    \item \textit{Language-guided reinforcement learning} is an emerging application investigating whether text descriptions of a task can help guide an agent to learn better policies in some environment~\citep{narasimhan2018grounding}. Unimodal encoders capture atomic units such as specific text or visual references to a single entity in the environment, the agent's possible actions, or possible ways of obtaining rewards. Learning correspondences refers to relating text descriptions to visual objects in the environment (e.g., identifying that text references to a weapon refer to images of weapons in the environment), actions in the action space, or possible rewards. Composing these references is crucial towards more efficiently learning a policy mapping visual states and text descriptions to a distribution over an agent's actions to maximize cumulative reward.
\end{enumerate}
We now describe these $3$ steps in detail and provide methodological examples suitable for commonly studied modalities including language, image, video, audio, and time-series data.

\vspace{-1mm}
\subsubsection{Encoders}
\label{sec:encoders}
\vspace{-1mm}

Each modality is first processed by a set of specialized encoders. The goal of each encoder is to capture unimodal syntax and semantics and summarize them into a unimodal representation. Unimodal syntax is captured by segmenting raw input data into a set of atomic units (for example, a sequence of words or word parts in the language modality, a set of object bounding boxes in the image modality, or a set of segmented speech parts in the audio modality). Unimodal semantics are captured by learning a feature vector summarizing the meaning of each atomic unit. In general, feature vectors learned by representation learning techniques exhibit certain desirable properties that make them suitable for capturing meaning~\citep{Bengio:2013:RLR:2498740.2498889}. We list some below but defer the reader to~\citet{Bengio:2013:RLR:2498740.2498889} for a more detailed treatment of the general desiderata of feature representations:
\begin{enumerate}
    \item \textit{Smoothness:} atomic units in each modality with similar meaning tend to be mapped into similar feature vectors in representation space.
    \item \textit{Multiple explanatory factors:} different underlying factors in the data generating distribution (e.g., size, shape, color for visual units) are encoded through different subspaces of the representation space.
    \item \textit{Hierarchy of explanatory factors:} underlying factors are defined in terms of other concepts in a hierarchy, with more abstract concepts higher in the hierarchy defined in terms of less abstract ones.
    \item \textit{Natural clustering:} different values of atomic units (e.g., object categories) cluster into separate manifolds in representation space, and local variations within each cluster tend to preserve the value of a category.
\end{enumerate}
To summarize, the output from each modality's encoder is a set of atomic features, each with information represented in a dense vector. By modeling unimodal syntax and semantics, the result is a fine-grained unimodal feature representation capturing both compositionality and meaning in unimodal data.

\textbf{Examples:} In practice, some common examples of machine learning encoders are convolutional neural networks~\citep{Lecun98gradient-basedlearning} for images, which have been shown to learn features representing the semantic meaning of the object in the image. More fine-grained models such as Region-CNNs~\citep{girshick2014rich} further extract feature representations of multiple objects in an image along with their bounding box regions. Another closely related line of models are those designed for image-based semantic segmentation~\citep{long2015fully} which categorizes every pixel in the image into a semantic object category.

For text and other sequential data such as speech and time-series, sequence models like Recurrent neural networks~\citep{rumelhart1986learning}, Long short-term memory networks~\citep{hochreiter1997long}, and Transformer models~\citep{vaswani2017attention} have emerged as the de-facto choice for processing. For discrete data like text, the set of discrete tokens is typically first converted into continuous space using a Tokenizer before learning a token embedding dictionary. For video data, methods for encoding images (such as convolutional networks or R-CNNs) are typically combined with a sequence model - the former image-based methods extract features for each frame in the video, and the sequence of features over all frames are combined with a sequence model such as a Long short-term memory network~\citep{hochreiter1997long} or Transformer network~\citep{vaswani2017attention}.

For graphs, graph-based neural networks have emerged as a popular option~\citep{scarselli2008graph,wu2020comprehensive}. Each unit is defined as a node or edge, and the combination function is determined by local connectivity within the graph structure. For example, representations of nodes would be combined if they were connected together by an edge (or a weighted combination in the case of weighted graphs).

For tables and sets, a commonly-adopted paradigm is to model the permutation-agnostic structure of input data using a permutation-agnostic model which has been shown to learn features that better respect the structure of input data~\citep{zaheer2017deep}.

\vspace{-1mm}
\subsubsection{Representation space}
\label{sec:representation}
\vspace{-1mm}

\begin{figure}[tbp]
\centering
\includegraphics[width=\linewidth]{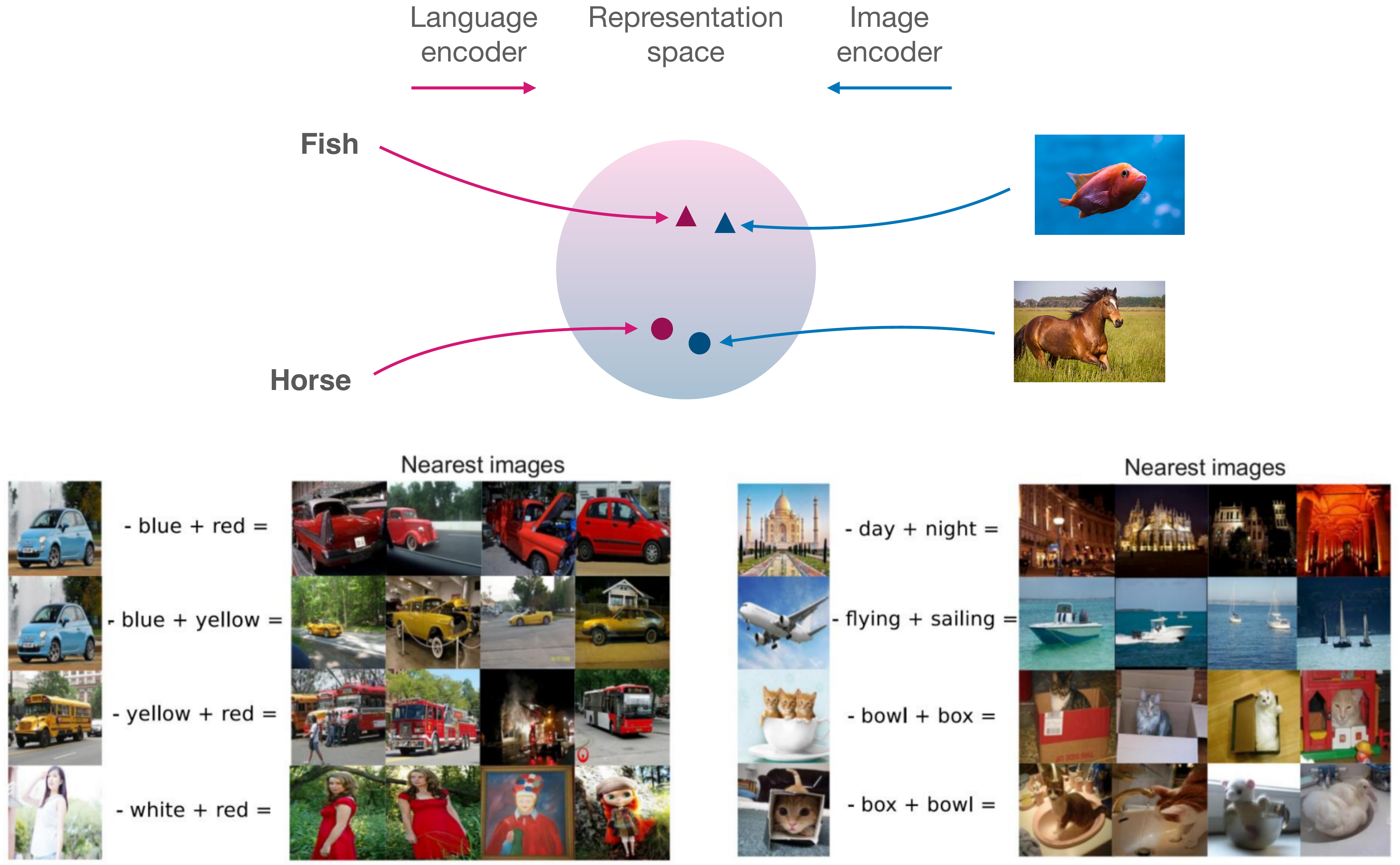}
\caption{Top: A general approach to learn a \textbf{coordinated representation space} is to define a set of atomic units that are known to correspond with each other (e.g., words with their corresponding images and sounds), and enforce alignment using an objective function that imposes structure in paired units. Bottom: A coordinated representation space enables fusion, retrieval, and compositionality in the form of multimodal vector space arithmetic (Figure from~\citep{kiros2014unifying}).}
\label{fig:space}
\vspace{-4mm}
\end{figure}

After learning atomic features in each modality, the core research problem lies in learning a representation space that takes in multiple unimodal feature representations and learns a coordinated multimodal representation capturing multimodal syntax and semantics. As formalized in Section~\ref{sec:properties}, multimodal semantics involves learning the correspondences/alignment in atomic features across modalities based on shared meaning. Multimodal syntax involves learning how aligned subsets of atomic features relate and compose with each other to derive holistic meaning across entire multimodal inputs (rather than at the level of units). The result is a coordinated multimodal representation capturing shared and composed meaning across multimodal inputs.

A general approach to learning this representation space is to define a set of atomic units that are known to correspond with each other, and enforce alignment using some objective function that imposes a certain form of structure in paired units (see Figure~\ref{fig:space}). Some typical examples of structure and their corresponding alignment objective functions are:

\textbf{Similarity measures} aim to learn a representation space containing transformed features from both modalities such that units of the same semantic meaning are mapped to features that are nearby in feature space. Distance is typically measured via cosine distance, l2 distance, or max-margin losses. The exact algorithm used to preserve semantic distances can range from contrastive learning~\citep{lu2019vilbert,yuan2021multimodal}, noise contrastive estimation~\citep{pielawski2020comir}, max-margin learning~\citep{guo2007enhanced}, or visual-semantic embedding models~\cite{frome2013devise}. When applied to images and captions, the resulting coordinated representation space enables compositional multimodal vector space: representation(image of blue car) - representation(word \textit{``blue''}) + representation(word \textit{``red''}) = representation(image of red car)~\cite{kiros2014unifying}.

\textbf{Ordered and hierarchical spaces:} In contrast to the above methods which are distance-preserving (semantically similar objects are mapped to points that are nearby in the embedding space), an alternative approach is to maintain an order-preserving representation space.~\citet{vendrov2015order} achieve this by constructing a visual-semantic hierarchy that captures a partial order of language and image representations. For example, the image of \textit{``a woman walking her dog''} should align with the text \textit{``woman walking her dog''} which falls under the text \textit{``woman walking''}, in order to better capture hierarchical representations of text and their subsentences.~\citet{zhang2016learning} also explore this idea in learning multimodal concept taxonomies resulting in a hierarchy of hypernyms (i.e., categorizing specific concepts with respect to more general ones) across both textual and visual atomic units.

\vspace{-1mm}
\subsubsection{Decoders}
\label{sec:decoders}
\vspace{-1mm}

Finally, the primary function of decoders is to map data in representation space into data space. We distinguish between 2 types of decoders:

1. \textit{Predictive decoders} map data in representation space into a set of labels for a particular task that one cares about (e.g., a set of object categories or human emotions). The prediction process typically takes in the set of aligned units across modalities and learns a task-specific composition function that combines these aligned features into a prediction for a task. For example, given corresponding atomic units (e.g., a paired positive word and loud voice), the predictive decoder would predict \textit{``strongly happy''} as the emotion displayed by the speaker in the video. Fused multimodal data in the form of coordinated representations is important to capture complementary information for prediction (e.g., being able to predict emotion when through only speaker gestures, or when language and speech are also present).

\textbf{Examples:} Predictive decoders are typically trained neural network classifiers where the composition function is approximated via gradient-based learning. Recent work has also explored handcrafting explicit composition functions~\citep{yi2018neural} based on the parse tree of questions for tasks like image-based question answering.

2. \textit{Generative decoders} map data in representation space back into high-dimensional data space such as that of images, natural language, or speech signals. Generation can be in the same modality (e.g., dialog prediction in language) or different modality (describing an image in language), all of which necessitate starting from a coordinated multimodal representation. A core challenge in generation is that of controllable interpolation - given a \textit{new feature sample} in representation space, can we decode it back into data space while respecting the changes in factors of variation in feature space? Factors of variation correspond to individual, atomic changes in the input modality. For example, given a visual object, each factor of variation could correspond to changes in its color, shape, size, and orientation~\citep{kulkarni2015deep}. Similarly, given a sentence, each factor of variation could correspond to changes in its tense, sentiment, and tone~\citep{lyu2021styleptb}. Controllable generation involves the ability to change each factor individually while achieving corresponding perceptible changes in the desired basis in output space, which is important for generating data with desired properties.

\textbf{Examples:} Recent work in generative decoders has formed the basis for much recent work in generative modeling of images, video prediction, and style transfer across the image, text, and video modalities. Some examples of decoders back to raw data include deconvolutional networks/upsampling for image generation~\cite{brock2018large} and autoregressive models for sequential data such as text, audio, and video~\citep{mehri2016samplernn,radford2019language,weissenborn2019scaling}.

\vspace{-1mm}
\subsection{Addressing the Technical Challenges}
\vspace{-1mm}

To see why this is a general framework for multimodal language learning, we explain how this approach is able to tackle each of the core technical challenges in multimodal machine learning as described in Section~\ref{sec:background}.

\textbf{Alignment:} The challenge of alignment is most directly tackled by this learning paradigm - alignment consists of the set of unimodal atomic units and their learned correspondences with atomic units in the same or different modality.

\textbf{Representation:} This framework provides flexibility in defining the representations at various levels. The first level is at the level of unimodal representations by representing each atomic unit as a feature. The second level is a multimodal representation defined as the composition of unimodal atomic units and their learned correspondences with atomic units in the same or different modality.

\textbf{Fusion} is performed when the overall multimodal representation from above is combined with a task-specific prediction layer to make a fused prediction (e.g., predicting speaker affect from human videos).

\textbf{Translation:} After aligning each unimodal atomic unit with their corresponding entities in the other modalities, translation from a source to target modality then amounts to retrieving from a set/generating from scratch the closest aligned unit in the target modality for each unit in the source modality. Special care still needs to be taken to coherently compose the retrieved/generated units into final raw data in the target modality (e.g., composing individual retrieved words/phrases into a full sentence).

\textbf{Co-learning} is indirectly exemplified by learning an alignment between modality representations. For example, learning that images of dogs (a visual unit) correspond to audio of dogs barking (an acoustic unit) induces shared information useful for both image classification as well as audio classification.

Therefore, our blueprint describing a general multimodal learning process is a step towards addressing several core technical challenges in multimodal learning.

\vspace{-2mm}
\section{Multimodal Language for Consciousness: A Case Study in the CTM}
\label{sec:ctm}
\vspace{-2mm}

Given the above treatment of a general multimodal language based on modality-specific encoders, a coordinated representation space, and modality/task-specific decoders, we explain how these can be applied in the Conscious Turing Machine (CTM), a machine model for consciousness as proposed by~\citet{blum2021theoretical}. \name\ comprising words, images, audio, and sensations combined in representations is an essential language that the CTM's processors use to communicate with each other, and enables higher-order cognitive functions such as multimodal processing, constructing a model of the world, inner speech, vision, and tactile sensation, as well as dreaming.
Please refer to Table~\ref{table:ctm} for an overview of CTM processors and their corresponding multimodal challenges.

\vspace{-1mm}
\subsection{Unimodal Processors}
\vspace{-1mm}

The unimodal processors that each process information from individual modalities can be seen as unimodal encoders that learn basic unimodal representations. Each of these representations contains information about unimodal atomic units, their feature representations, and how atomic units are structured compositionally (essentially unimodal syntax and semantics as outlined in Section~\ref{sec:properties}). Each unimodal processor can take the form of a unimodal encoder as described in Section~\ref{sec:encoders}.

\vspace{-1mm}
\subsection{\name\ Multimodal Language}
\vspace{-1mm}

The \name\ multimodal language is defined as the \textit{coordinated representation space} across all sensory modalities that have been processed by unimodal processors, and multimodal features that have been extracted by multimodal processors. These features are summarized as multimodal gists~\citep{blum2021theoretical} in the coordinated representation space as described in Section~\ref{sec:representation}. The \name\ multimodal language can then be used for multimodal processors in the CTM such as audio-visual fusion, emotion recognition, model-of-the-world, inner speech/vision/sensation, and dreaming which we will describe in the next subsection.

\vspace{-1mm}
\subsection{Multimodal Processors}
\vspace{-1mm}

\begin{table*}[]
\fontsize{9}{11}\selectfont
\setlength\tabcolsep{3.0pt}
\vspace{-4mm}
\caption{Linking core processors in the CTM with multimodal research challenges.}
\centering
\footnotesize
\vspace{-0mm}
\begin{tabular}{c|cccccccc}
\hline \hline
\multicolumn{1}{c|}{CTM processor} & Examples & Challenge \\
\hline
\multirow{1}{*}{Unimodal processors} & Language, vision, audio, smell, touch & Unimodal representation & \\
\hline
\multirow{6}{*}{Multimodal processors} & Audio-visual & Multimodal fusion\\
& Language-visual & Multimodal fusion \\
& Emotion & Multimodal fusion \\
& Model-of-the-world & Multimodal fusion \\
& Inner speech, vision, sensation & Multimodal translation & \\
& Dreams & Multimodal generation & \\
\hline
\multirow{1}{*}{CTM language} & Brainish & Multimodal representation \\
\hline \hline
\end{tabular}
\vspace{-4mm}
\label{table:ctm}
\end{table*}

Multimodal processors integrate information from representations learned by individual unimodal processors to summarize multimodal information. Based on our general treatment of multimodal learning, each multimodal processor works by learning the correspondences between atomic units across modalities and composing these to form multimodal features useful for multimodal tasks. We outline some examples of multimodal processors below:

\textbf{Audio-visual processor:} The McGurk effect~\cite{mcgurk1976hearing} shows that the brain processes information from audio and visual sensory inputs in order to recognize speech from a speaker. Multimodal learning exists to the extent that when these inputs conflict with each other (ambiguity), an `overriding' phenomenon occurs where misreading the person's lips leads to incorrect inferences on predicted speech. This could be realized by an audio-visual processor that learns and composes the correspondences between mouth movements and auditory features (i.e., \textit{``baa baa baa'', ``daa daa daa''}) into order to make a prediction of the spoken speech~\citep{ngiam2011multimodal}.

\textbf{Language-visual processor:} Integration of language and vision is crucial for human cognition since language is commonly used as a communicative medium in reference to the visual world. The integration of language and vision is exemplified through machine learning tasks such as question answering (asking a question in reference to an image/video and obtaining a correct answer)~\citep{andreas2016neural,vedantam2019probabilistic,yi2018neural}, navigation (giving a text instruction in reference to a visual environment and obtaining a sequence of steps taken in that environment to complete the instruction~\citep{majumdar2020improving}), and image captioning (generating relevant text descriptions of a given image~\citep{vinyals2016show}).

\textbf{Emotion processor:} An emotion processor aims to recognize human sentiment and emotion through multimodal communicative behaviors spanning language (spoken words), visual (facial expressions, gestures), and acoustic (prosody, speech tone)~\cite{liang2018computational}. These processors have also been studied in machine learning literature, and several strong-performing methods are primarily based on the idea of multimodal temporal fusion of these heterogeneous signals~\citep{zadeh2018memory}.

\begin{figure}[tbp]
\centering
\includegraphics[width=\linewidth]{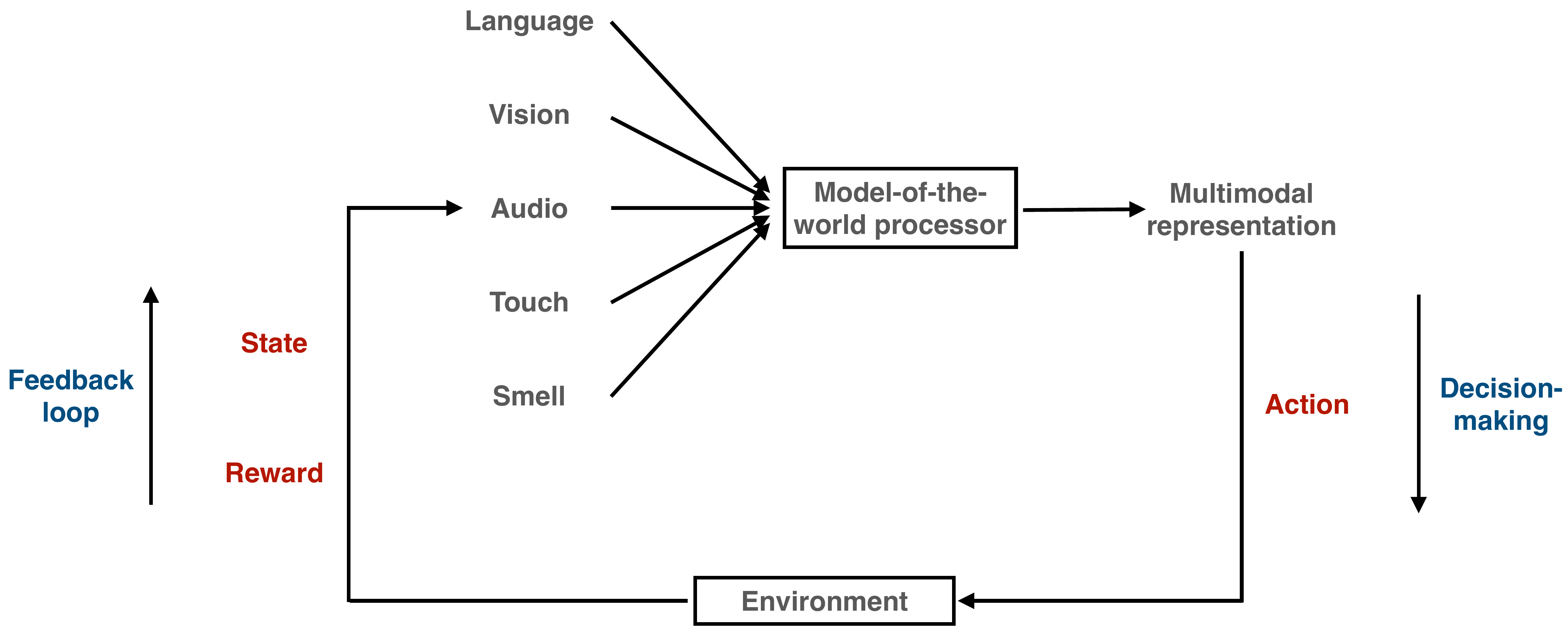}
\caption{\textbf{The model-of-the-world processor} is an example of a multimodal processor which combines the agent's multisensory information observed from the physical world into a multimodal representation (gist), plans possible actions in the world, and updates itself based on the effect that its actions have on the world (i.e., by observing new states and rewards as an outcome of its actions).}
\label{fig:mow}
\vspace{-4mm}
\end{figure}

\textbf{Model-of-the-world processor:} Another important multimodal processor in the CTM is the agent's model-of-the-world processor which constructs models of its inner and outer worlds. These multimodal models of the world combine the agent's multisensory information observed from the physical world, plan possible actions in the world, predict the effect that its actions have on the world, and help distinguish self from not-self~\citep{blum2021theoretical} (see Figure~\ref{fig:mow}). These models are crucial for recognizing strange settings, planning actions by modeling their impact on the environment, and helping the CTM to stay out of danger. Having a multimodal language is a crucial component in meeting each of these goals.

Prior research in combining multisensory information from the physical world has been studied in multisensor fusion (involving, for example, the visual, audio, and force modalities) for robotics~\citep{lee2019making,liang2021multibench,savva2019habitat}. In these settings, the agent's policies (a mapping from multimodal inputs to actions) are learned based on fused multimodal representations~\citep{chaplot2017gated}. Multimodal representations are essential for good action prediction since each modality provides unique information not present in the others. Furthermore, multimodal fusion enables one to perform prediction in the face of noisy or missing modalities, such as relying on the visual modality to predict robotic movement when touch sensors are missing and vice versa~\citep{lee2019making}. Furthermore, predicting the effect that an agent's actions have on the world can be seen as a form of model-based reinforcement learning, which again has been extensively applied to multimodal settings~\citep{luketina2019survey}.

\textbf{Inner speech, vision, and tactile sensation:} The inner speech processor takes a gist representation broadcast by the CTM's short-term memory (i.e., its stage) and maps it to the same locations that the input sends representations of outer speech~\cite{blum2021theoretical}, and can be nearly indistinguishable from the gists of actual speech from the environment~\cite{rosen2018tangled}. This is a \textit{multimodal translation} problem that involves taking in a representation of one input modality and decoding it into (inner) speech, which is exemplified by similar machine learning tasks such as conditional text generation~\citep{openai} (inner speech given read text), conditional image/video captioning~\citep{lin2014microsoft,rohrbach2017movie} (inner speech given observed visual scenes), and multimodal dialog~\citep{nie2019multimodal} (inner speech given heard dialog possibly paired with observed visual scenes). Again, it is important that the gist representation obtained from an observed input is mapped into the \name\ multimodal space to enable translation from an arbitrary source modality to output speech. Without coordination, translation would not be possible from independently-learned representation spaces of source modalities (i.e., visual) and target modalities (e.g., speech). Similarly, the generalized inner speech processors for inner vision and inner sensation~\cite{blum2021theoretical} are also translation problems with different output modalities and require a multimodal gist representation to enable semantically-aligned decoding into arbitrary outputs. Decoding into the desired target modality can be performed by any of the decoders as described in Section~\ref{sec:decoders}.

\textbf{Dreams} demonstrate the power of Brainish gists: what the CTM sees, hears, feels, and does in a dream are fabrications by processors that can recall, modify, and submit creations to the competition for short-term memory~\cite{blum2021theory}. Dreams generate the sense of a realistic world even while the CTM is completely divorced from external inputs, and can appear so realistic that it may become hard to distinguish dreams from reality~\cite{corlett2014dreams}.
Dreaming can therefore be seen as taking a semantic representation in the form of gists stored in memory~\cite{zadra2021brains} and decoding it into long-range multimodal data with no feedback from the outside world during this generation process.
Dreaming can be viewed as a \textit{multimodal generation} problem where semantically meaningful mappings are learned from the gist representation to a series of long-range parallel modalities (which could span auditory, tactile, gustatory, and olfactory dream components~\cite{meaidi2014sensory}). This decoding process is often (1) \textit{conditional}, (2) \textit{synchronized}, (3) \textit{stochastic}, and (4) \textit{auto-regressive}. Dreams don't replay memories exactly, but are semantically conditioned on the same gist as some recent memory and could have the same title~\cite{zadra2021brains}. It is synchronized across modalities since dreams involve output modalities that are semantically coherent. The process is stochastic since there are many possible future generations given a particular state. Finally, it is auto-regressive across possibly long ranges: future dream states are recursively generated given previous ones. In practice, there has been some progress towards text to image generation~\citep{ramesh2021zero}, text to speech~\cite{taylor2009text}, image captioning~\cite{vinyals2016show}, and video generation~\citep{oh2015action}, but a complete decoding process into synchronized high-dimensional multimodal data still remains a challenge for modern machine learning methods.

\vspace{-1mm}
\subsection{Main Take-away Messages}
\vspace{-1mm}

From this case study of a multimodal language in the Conscious Turing Machine (CTM)~\citep{blum2021theoretical}, we observe that many of the core cognitive processors require a multimodal language to function. These span processors whose main purpose is to perform multimodal fusion (e.g., audio-visual reasoning, emotion recognition, model-of-the-world), multimodal translation (inner speech, vision, and sensation), and multimodal generation (dreams). The \name\ multimodal language is used by these processors to communicate with each other. Achieving a real-life implementation of such computational models therefore requires understanding the functionality of a multimodal language simultaneously realizing multimodal fusion, translation, and generation.

\vspace{-2mm}
\section{Multimodal Language for Intelligence: A Case Study in Machine Learning}
\label{sec:ml}
\vspace{-2mm}

\begin{figure*}[tbp]
\centering
\vspace{-0mm}
\includegraphics[width=\linewidth]{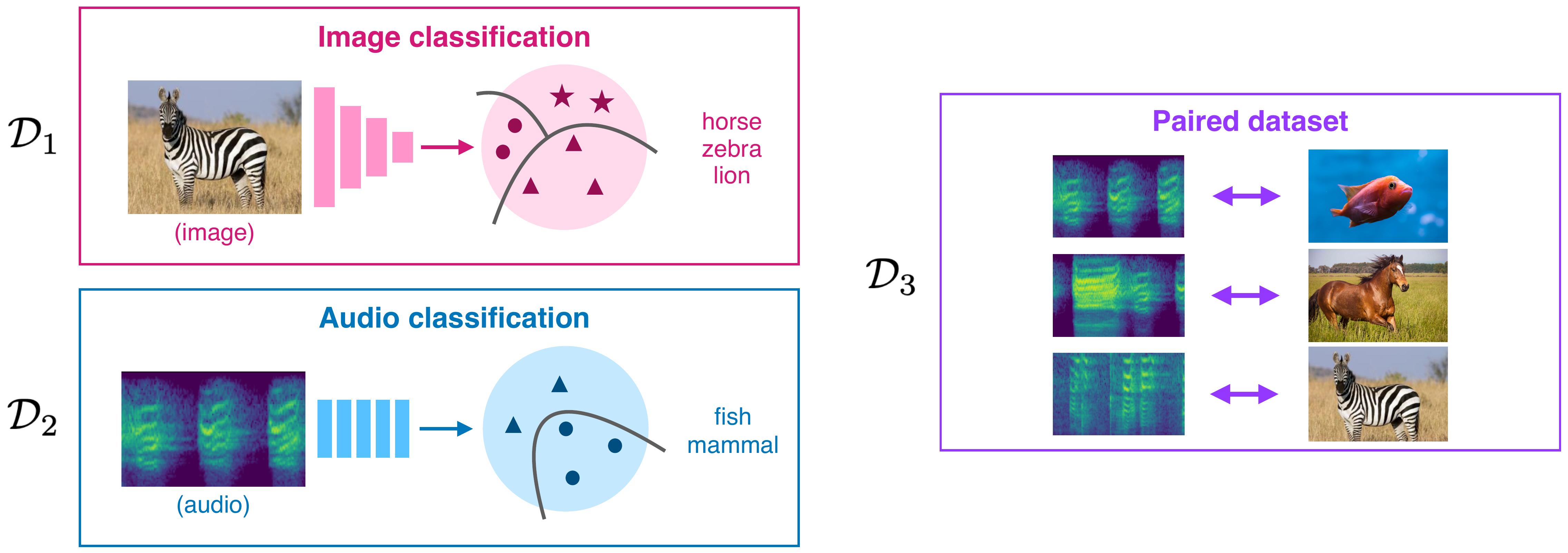}
\vspace{-4mm}
\caption{Experimental setup for a real-world implementation of multimodal language with $2$ modalities. $2$ datasets each labeled for some prediction label $\mathcal{D}_1 = \{ (x_1, y)\}$ and $\mathcal{D}_2 = \{ (x_2, y)\}$ represent feedback we obtain separately for images and audio. The paired dataset $\mathcal{D}_3 = \{ (x_1, x_2)\}$ represents shared information through paired natural occurrences of images and audio~\cite{liang2020cross}.\vspace{-2mm}}
\label{ml}
\end{figure*}

In this section, we describe a real-world implementation of multimodal language in the context of a problem with $2$ modalities. Specifically, the setup, which we illustrate in Figure~\ref{ml}, consists of $2$ datasets each labeled for some prediction label $\mathcal{D}_1 = \{ (x_1, y)\}$ and $\mathcal{D}_2 = \{ (x_2, y)\}$ as well as a paired dataset $\mathcal{D}_3 = \{ (x_1, x_2)\}$ across modalities. This general setup captures a natural scenario in both human and artificial intelligence where feedback (the label) is provided for either modality - for example, feedback is provided for images through dataset $\mathcal{D}_1$ and feedback is provided separately for audio through dataset $\mathcal{D}_2$. To enable multimodal learning, one also needs some amount of paired data across both modalities (e.g., paired natural occurrences of images and audio) through $\mathcal{D}_3$~\cite{liang2020cross}. Note that this setup does not require labels for paired data across both modalities (i.e., feedback for paired image and audio at the same time).

\vspace{-1mm}
\subsection{Datasets and Tasks}
\vspace{-1mm}

We collect the following datasets across the following paired modalities: text and image, image and audio, as well as text and speech. Code for data, methods, and experiments in this section can be found at \url{https://github.com/pliang279/Brainish/}.

\textbf{Text and image dataset:} We use the Yummly-28K dataset~\cite{yummly_dataset} which contains parallel text descriptions and images of recipes. We create classification labels from the metadata by concatenating the meal-type and cuisine, yielding $44$ distinct classes. A large number of recipes and shared concepts between text and image makes it an ideal testbed for learning a shared multimodal language.

\textbf{Image and audio dataset:} We combine two large unimodal classification datasets over images (CIFAR-$10$ and CIFAR-$100$~\cite{krizhevsky2009learning}) and audio made by various objects (ESC-$50$~\cite{piczak2015esc}) with partially related label spaces. This allows us to leverage complementary information from both modalities while testing on new concepts. To obtain weak pairs, we map similar classes between the datasets using similarities from WordNet~\cite{Miller:1995:WLD:219717.219748} and text cooccurrence. This yields $17$ clusters of weak pairs.

\vspace{-1mm}
\subsection{Learning a Multimodal Language}
\vspace{-1mm}

We now describe our approach of learning a multimodal language (\name\ for short) on the union of these $3$ datasets.

\textbf{Encoders:} We define encoders $e_s, e_t$ for source and target modalities (i.e., one specialized encoder for each modality). Specifically, the encoders we use are ResNet pretrained on ImageNet~\cite{deng2009imagenet} to encode the images, pretrained BERT encoder~\cite{devlin2018bert} for text, and a Convolutional neural network (CNNs) pretrained on AudioSet~\cite{gemmeke2017audio} to encode audio~\cite{kumar2018knowledge,ridnik2020tresnet}. ResNets are designed for image processing with an inductive bias inspired by convolutional layers which model spatial locality in images. and have shown state-of-the-art results in image classification. BERT is a recent model for processing text which takes into account both features of individual words as well as how they are used in bidirectional context: how the meaning of words is influenced by the meaning and order of words before and after in a sentence. Convolutional neural networks are strong models for processing audio spectrograms by treating spectrograms as an image waveform.

\textbf{Representation space:} We aim to learn a representation space that takes in multiple unimodal feature representations and learns a coordinated multimodal representation capturing multimodal syntax and semantics. Multimodal semantics involves learning the correspondences/alignment in atomic features across modalities based on shared meaning. Multimodal syntax involves learning how aligned subsets of atomic features relate and compose with each other to derive holistic meaning across entire multimodal inputs. To achieve this, we use dataset $\mathcal{D}_3$ which contains pairs across modalities of the form $(x_1, x_2)$. We define a similarity measure such that units of the same semantic meaning, as represented by paired units $(x_1, x_2)$, are mapped to features that are nearby in feature space (i.e., alignment).

We model alignment by learning an alignment function $p (a|x_1,x_2)$ which outputs a probability $a$ representing the likelihood of $x_1$ and $x_2$ being semantically matched. We parametrize $p(a|x_1,x_2) \propto e_1(x_1)^\top e_2(x_2)$ which is a natural way of measuring (unnormalized) similarity based on cosine similarity of vectors $e_1(x_1)$ and $e_2(x_2)$ in the coordinated representation space. Training requires positive samples $(x_1, x_2) \in \mathcal{D}_3$ for which we would like to maximize $p(a|x_1,x_2)$, but also requires contrastive negative samples $x_1, x_{2, \textrm{neg}}$ sampled randomly across all pairs in $\mathcal{D}_3$ for which we would like to minimize $p_\theta(a|x_1,x_2)$. The overall objective resembles Noise Contrastive Estimation (NCE)~\cite{dyer2014notes} which learns a binary classifier to distinguish paired samples $(x_1,x_2) \in \mathcal{D}_3$ from unpaired negative samples $x_{2, \textrm{neg}}$.

Therefore, given encoders $e_s, e_t$ for source and target modalities and paired dataset $\mathcal{D}_3$, we learn an aligned space across source and target modalities by optimizing for the NCE loss:
\begin{align}
    \mathcal{L}_\textrm{align} &= \sum_{(x_1,x_2) \in \mathcal{D}_3} \left( - \log p(a|x_1,x_2) + \sum_{x_{2, \textrm{neg}}} \log p(a|x_1,x_{2, \textrm{neg}}) \right) \\
    &\propto \sum_{(x_1,x_2) \in \mathcal{D}_3} \left( - e_1(x_1)^\top e_2(x_2) + \sum_{x_{2, \textrm{neg}}} e_1(x_1)^\top e_2(x_{2, \textrm{neg}}) \right).
    \label{align_eq}
\end{align}
where $x_{2, \textrm{neg}}$ denotes unpaired negative samples. The NCE objective has a nice interpretation as capturing a space where the representations of similar concepts expressed in different modalities are close together, and different concepts are far apart~\cite{frome2013devise,DBLP:journals/corr/abs-1806-03560}.

\textbf{Decoders:} Given an aligned space, we now train a single classifier $\phi$ on top of the aligned space for prediction across datasets $\mathcal{D}_1 = \{ (x_1, y)\}$ and $\mathcal{D}_2 = \{ (x_2, y)\}$ by optimizing for the cross-entropy loss, which maximizes the log probability of predicting the true label $y$ given data $x_1$ (or $x_2$):

\begin{align}
    \mathcal{L}_1 = \sum_{(x_1,y) \in \mathcal{D}_1} - \log \phi (y | e_1 (x_1)), \qquad \mathcal{L}_2 = \sum_{(x_2,y) \in \mathcal{D}_2} - \log \phi (y | e_2 (x_2)).
    \label{loss_eq}
\end{align}

\begin{figure}
    \begin{minipage}{\linewidth}
    \vspace{-2mm}
    \begin{algorithm}[H]
    \caption{Learning a multimodal language across $2$ modalities.}
    \begin{algorithmic}
    \State Initialize encoders $e_1$ and $e_2$, classifier $\phi$.
    \For{iteration = $1,2,\dots$}
        \State Sample alignment pairs $\{ x_1, x_2 \}$ from dataset $\mathcal{D}_3$.
        \State Compute alignment loss $\mathcal{L}_\textrm{align}$ (equation~\ref{align_eq}) on pairs $\{ x_1, x_2 \}$.
        \State Update $e_1 := e_1 - \nabla_{e_1} \mathcal{L}_\textrm{align}$ and $e_2 := e_2 - \nabla_{e_2} \mathcal{L}_\textrm{align}$ using gradient updates.
        \State Sample modality 1 pairs $\{ x_1, y \}$ from dataset $\mathcal{D}_1$.
        \State Compute prediction loss $\mathcal{L}_1$ (equation~\ref{loss_eq}) on pairs $\{ x_1, y \}$.
        \State Update $e_1 := e_1 - \nabla_{e_1} \mathcal{L}_1$ and $\phi := \phi - \nabla_{\phi} \mathcal{L}_1$ using gradient updates.
        \State Sample modality 2 pairs $\{ x_2, y \}$ from dataset $\mathcal{D}_2$.
        \State Compute prediction loss $\mathcal{L}_2$ (equation~\ref{loss_eq}) on pairs $\{ x_2, y \}$.
        \State Update $e_2 := e_2 - \nabla_{e_2} \mathcal{L}_2$ and $\phi := \phi - \nabla_{\phi} \mathcal{L}_2$ using gradient updates.
    \EndFor
    \end{algorithmic}
    \label{algotex}
    \end{algorithm}
    \end{minipage}
    \vspace{-2mm}
\end{figure}

\begin{figure*}[h]
\centering
\includegraphics[width=0.6\linewidth]{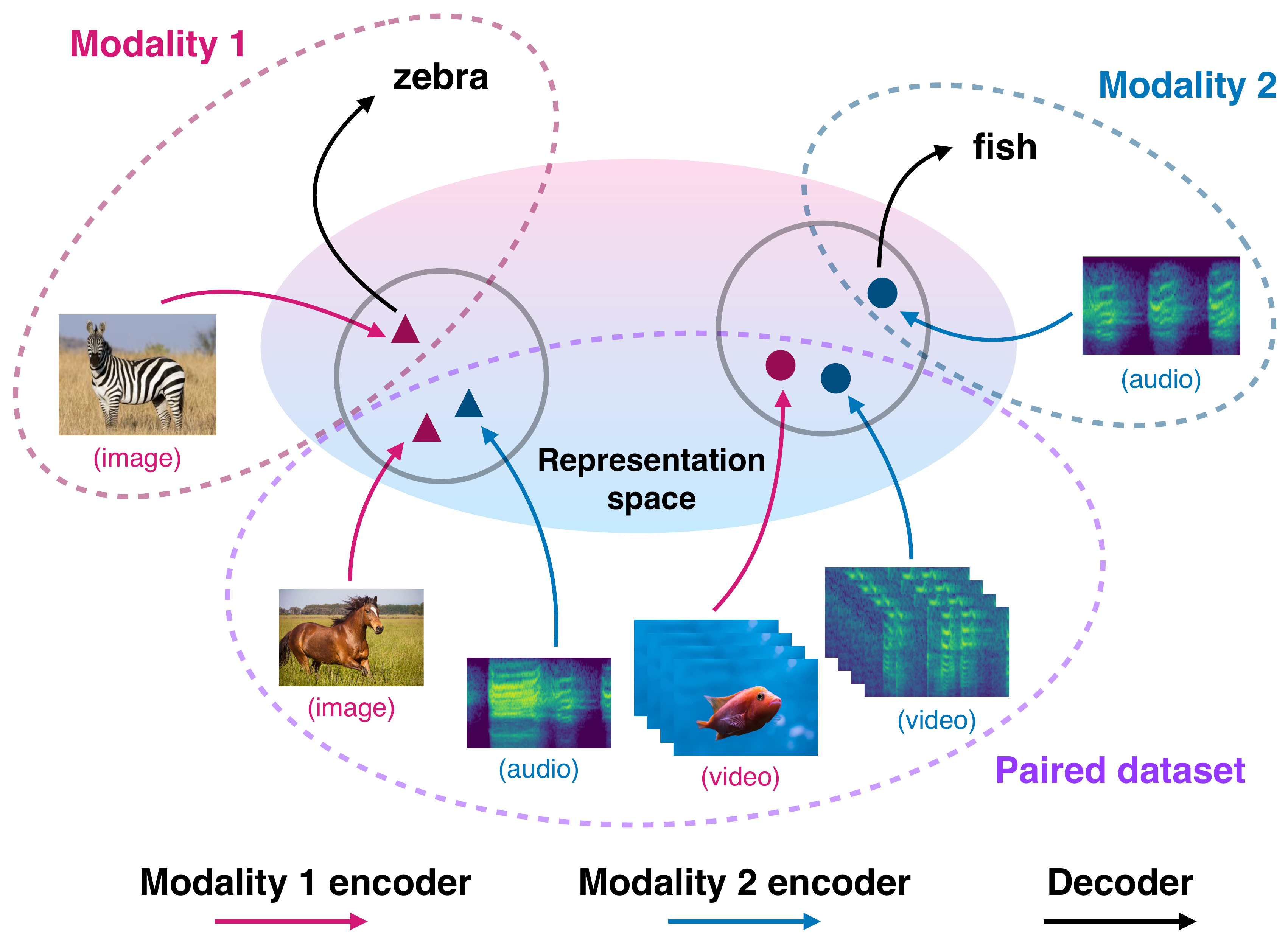}
\vspace{-0mm}
\caption{Visual depiction of learning a multimodal language. \textit{Encoders} take in raw data from different modalities and model unimodal syntax and semantics. Syntax is captured by segmenting each modality into atomic units, and semantics are captured by learning a representation summarizing the meaning of each unit. The \textit{representation space} takes in features across modalities and captures multimodal syntax and semantics. Multimodal semantics are learned via alignment: the matching across multiple modalities based on shared meaning. Multimodal syntax involves learning how aligned subsets of atomic units compose to derive holistic meaning across entire multimodal inputs. Finally, \textit{decoders} take in multimodal representations and output a prediction. \vspace{-2mm}}
\label{algo}
\end{figure*}

\textbf{Training and testing:} Overall, the training stage consists of learning from the alignment dataset $\mathcal{D}_3$ as well as classification tasks from each modality through datasets $\mathcal{D}_1$ and $\mathcal{D}_2$. We show the full training algorithm in Algorithm~\ref{algotex} and a visual diagram in Figure~\ref{algo}.

After training, we obtain trained encoder parameters $e_1, e_2$ and a classification decoder $\phi$. Given new data from $x_1$ (or $x_2$), a prediction is made by computing $\phi ( e_1 (x_1))$ or $\phi ( e_2 (x_2))$. In addition to prediction, this multimodal language also enables multimodal translation. Given data $x_1$, encode it into the coordinated representation space $e_1 (x_1)$ and rank its alignment with a samples $x_2$ in modality 2 by $x_2 = \argmax_{x_2} e_1(x_1)^\top e_2(x_2)$, which retrieves the most semantically aligned data from modality 2 matching input $x_1$.

\vspace{-1mm}
\subsection{Experiments}
\vspace{-1mm}

\begin{figure}[tbp]
\centering
\vspace{-0mm}
\includegraphics[width=\linewidth]{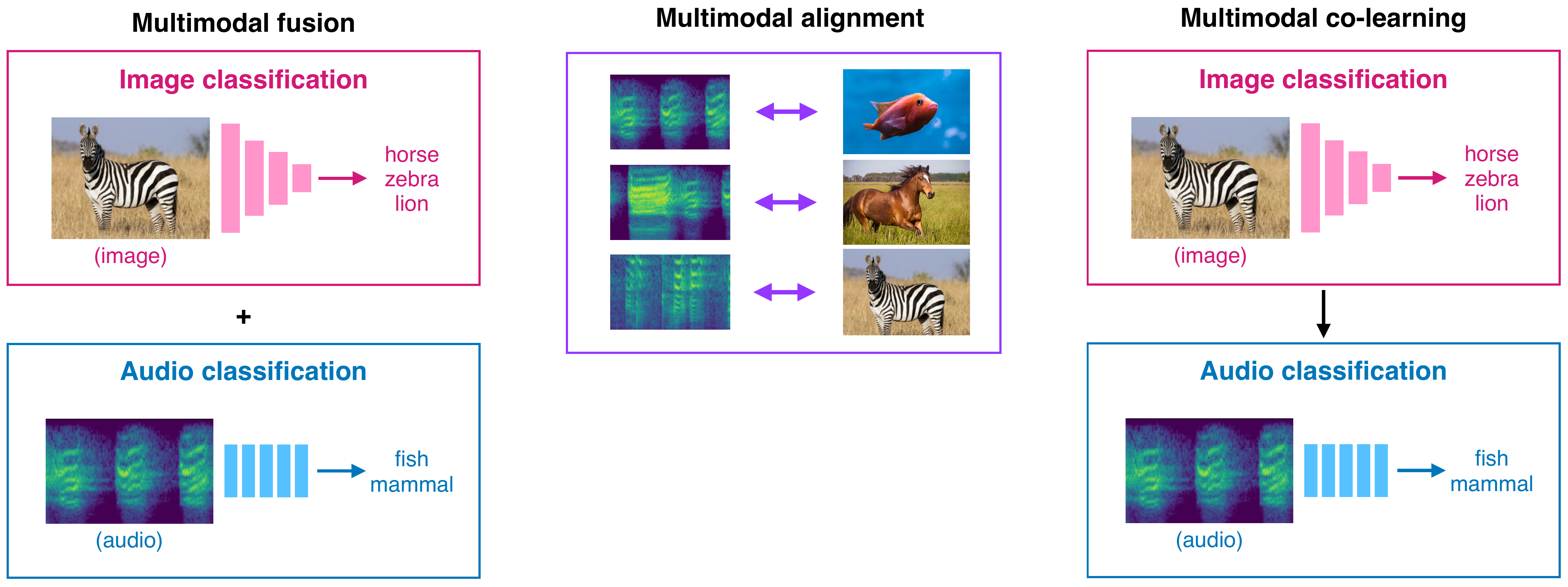}
\vspace{-4mm}
\caption{We design $3$ experimental settings to evaluate multimodal language learning. (1) In \textit{multimodal fusion}, we investigate whether a joint model learned from both image and audio classification tasks improve over separate models trained on each task alone. (2) In \textit{multimodal alignment}, we investigate whether the joint model can retrieve semantically similar data across modalities. (3) In \textit{multimodal co-learning}, we investigate whether it is possible transfer knowledge learned from one modality (image) to help a computational model trained on a different modality (audio).\vspace{-2mm}}
\label{setup}
\end{figure}

We design $3$ experimental settings to evaluate multimodal language learning across a suite of technical challenges described in Section~\ref{sec:background}. These settings are (1) multimodal fusion, (2) multimodal alignment, and (3) multimodal co-learning (see Figure~\ref{setup}).

\subsubsection{Experiment 1: Multimodal fusion}

\textbf{Setup:} We investigate whether a joint model learned from both image and audio classification tasks improves over separate models trained on each task alone. The former is our joint multimodal language model while the latter is a unimodal baseline that trains separate encoders $e_1, e_2$ and separate classifiers $\phi_1, \phi_2$ without sharing a common representation space. This baseline performs learning and prediction separately in each modality. We report classification accuracy in each modality, repeating experiments $10$ times to report mean and standard deviations.

\begin{table*}[t]
\fontsize{9}{11}\selectfont
\centering
\caption{Results on multimodal fusion across text and image classification. \name\ outperforms unimodal baselines that do not learn a multimodal language.}
\vspace{-0mm}
\setlength\tabcolsep{2.0pt}
\begin{tabular}{C{1.7cm} L{3cm} || *{3}{C{1.6cm}} C{2.7cm}}
\hline \hline
{\sc Task} & {\sc Approach} & {\sc Accuracy} \\
\Xhline{0.5\arrayrulewidth}
\multirow{3}{*}{\shortstack{{\color{gg}Text}\\+\\{\color{rr}Image}}} & \multirow{1}{*}{Unimodal text} & $60.0$ \\
\cline{2-6}
& \multirow{1}{*}{Unimodal image} & $55.0$ \\
\cline{2-6}
& \multirow{1}{*}{\name\ (ours)} & $\mathbf{68.0}$ \\
\hline \hline
\end{tabular}

\label{fusion}
\vspace{0mm}
\end{table*}

\begin{figure}[tbp]
\centering
\vspace{-0mm}
\includegraphics[width=0.8\linewidth]{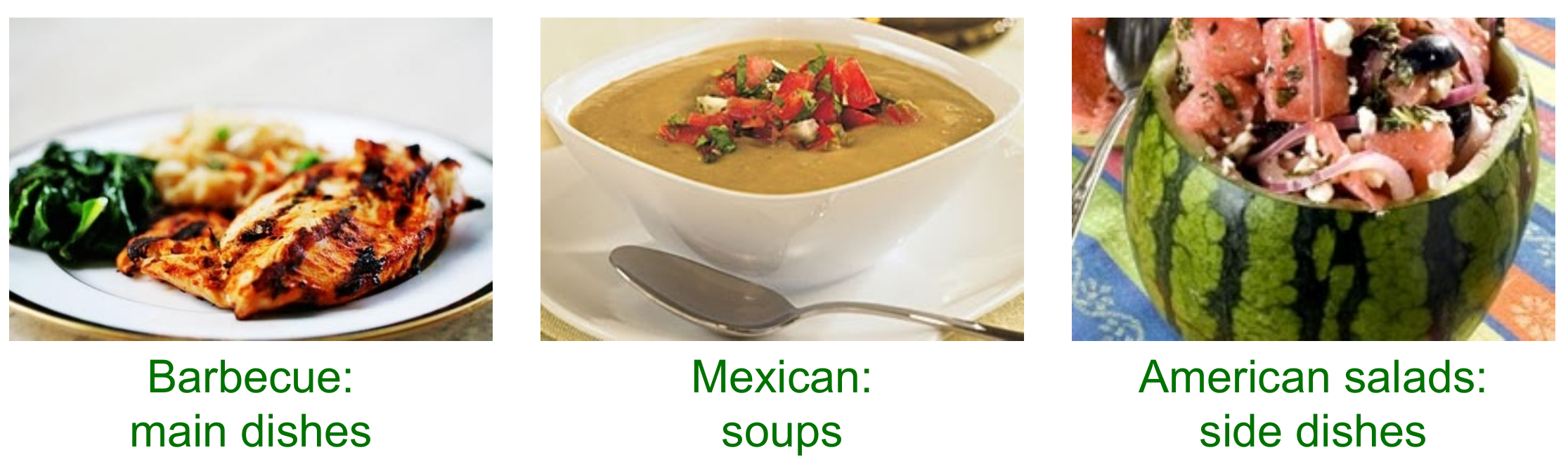}
\vspace{-0mm}
\caption{On Yummly-28K dataset, \name\ leverages source text to make accurate few-shot predictions on target images despite only seeing $1-10$ labeled image examples.\vspace{-2mm}}
\label{examples}
\end{figure}

\textbf{Results:} From Table~\ref{fusion} on text and image classification, \name\ outperforms unimodal approaches. This implies that discovering common information across both modalities through learning a multimodal language leads to performance gains over unimodal learning. Similar observations were also made in the field of multimodal fusion where multiple complementary signals improve performance in a variety of applications such as healthcare, robotics, multimedia, and affective computing~\cite{liang2021multibench}. We show samples of text and image classification into recipes in Figure~\ref{examples}. Our method is able to quickly recognize images from new recipes.

\subsubsection{Experiment 2: Multimodal alignment}

\textbf{Setup:} Our second experiment centers on the accuracy of multimodal alignment: given new data in one modality, does our approach accurately retrieve semantically-corresponding data in the other modality? Retrieval is measured using recall$@k$, rank, and cosine loss metrics~\cite{frome2013devise} with respect to the ground truth pairings in a held-out test set of $\mathcal{D}_3 = \{ (x_1, x_2)\}$.

\begin{figure*}[tbp]
\centering
\includegraphics[width=0.8\linewidth]{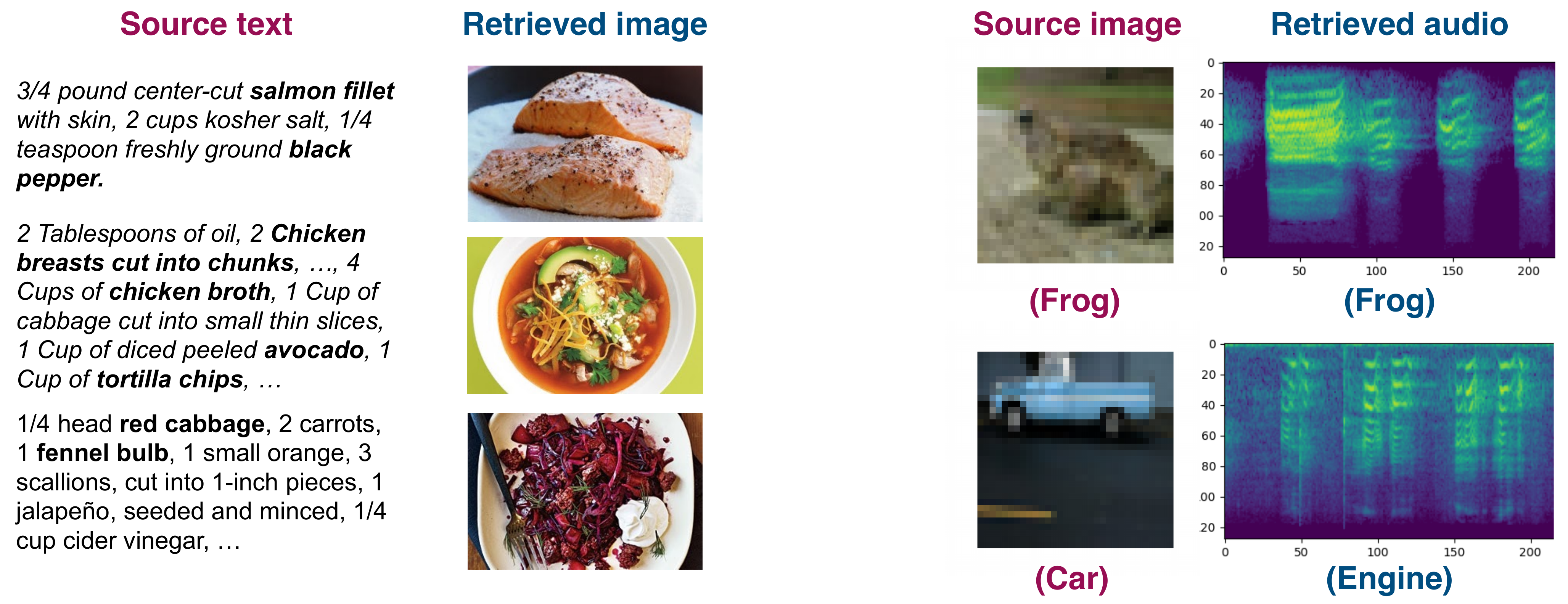}
\vspace{-0mm}
\caption{\textbf{Left}: samples of retrieved images given text recipes. \textbf{Right}: samples of retrieved audio samples given images. \name\ can perform multimodal retrieval at fine granularities.\vspace{-2mm}}
\label{pairs}
\end{figure*}

\textbf{Results:} We show retrieval performance in Table~\ref{alignment}. Our model yields better retrieval performance than a baseline that does not perform alignment of representation space, which indicates that alignment successfully aligns concepts across modalities to enable multimodal alignment. In Figure~\ref{pairs}, we also show samples of retrieved data in the target given input in the source modality to help us understand which source modalities the model is basing its target predictions on. We observe that the multimodal language is able to perform alignment at fine granularities.

\definecolor{gg}{RGB}{15,125,15}
\definecolor{rr}{RGB}{190,45,45}

\begin{table*}[t!]
\vspace{-0mm}
\centering
\fontsize{9}{11}\selectfont
\caption{Results on multimodal alignment: \name\ yields better alignment scores than the baselines, indicating that meta-alignment can align new concepts using only \textit{weakly paired data} across image and audio.}
\vspace{-0mm}
\setlength\tabcolsep{3.0pt}
\begin{tabular}{l l | c c c c c}
\hline \hline
{\sc $K$} & {\sc Approach} & {\sc R$@1$} \textcolor{gg}{$\uparrow$} & {\sc R$@5$} \textcolor{gg}{$\uparrow$} & {\sc R$@10$} \textcolor{gg}{$\uparrow$} & {\sc Rank} \textcolor{gg}{$\downarrow$} & {\sc Cos.} \textcolor{gg}{$\downarrow$} \\
\Xhline{0.5\arrayrulewidth}
\multirow{2}{*}{$5$} & No alignment & $1.0\%$ & $2.0\%$ & $5.5\%$ & $101$ & $0.428$\\
& \name\ (ours) & $\mathbf{4.0\%}$ & $\mathbf{19.5\%}$ & $\mathbf{39.0\%}$ & $\mathbf{13}$ & $\mathbf{0.003}$\\
\Xhline{0.5\arrayrulewidth}
\multirow{2}{*}{$10$} & No alignment & $0.5\%$ & $3.0\%$ & $4.5\%$ & $101$ & $0.399$\\
& \name\ (ours) & $\mathbf{3.5\%}$ & $\mathbf{17.5\%}$ & $\mathbf{35.0\%}$ & $\mathbf{15}$ & $\mathbf{0.004}$\\
\hline \hline
\end{tabular}
\label{alignment}
\vspace{-2mm}
\end{table*}

\subsubsection{Experiment 3: Multimodal co-learning}

\textbf{Setup:} Finally, our third experiment investigates multimodal transfer (co-learning): whether it is possible to transfer knowledge learned from one modality (e.g., predicted labels or representation) to help a computational model trained on a different modality? This challenge is particularly relevant when one of the modalities has limited resources~\cite{liang2020cross}. Using the same datasets, we study the transfer of knowledge from text to image, image to audio, and text to speech classification. We call the first task the source modality and the second task the target modality, which often is a low-resource modality with less labeled data than the source.

During training, we first learn classification in the source task (training $e_1$ and $\phi$) and alignment across source and target tasks (training $e_1$ and $e_2$). After training, we transfer the learned model to the target task. Using only a small number ($k$) of labeled training datapoints in the target, we update the model to perform target task classification (training $e_2$ and $\phi$). Using only $k$ labeled training datapoints in the target enables us to simulate few-shot learning settings under limited labeled target modality data, and truly test the capabilities of knowledge transfer from source to target modalities~\citep{finn2017model,NIPS2019_8731}.

We compare to the following baselines:
\begin{enumerate}
    \item Unimodal, which directly performs target task classification with $k$ labeled training datapoints in the target, without using information from the source task.
    \item Oracle, which performs target task classification with all labeled training datapoints in the target, which gives an upper bound on performance when there is no limited data in the target.
\end{enumerate}

\begin{table*}[t]
\fontsize{9}{11}\selectfont
\centering
\caption{Performance on multimodal co-learning: transferring knowledge from a source to target modality - text to image classification (top), and image to audio concept classification (bottom). \name\ is on par and sometimes outperforms the oracle target modality classifier that has seen thousands of labeled target samples, and also outperforms unimodal baselines that do not learn a multimodal language. \#Target (labels) denotes the number of target modality samples and labels used during meta-training.}
\vspace{-0mm}
\setlength\tabcolsep{2.0pt}
\begin{tabular}{C{1.7cm} L{3cm} | *{3}{C{1.6cm}} C{2.7cm}}
\hline \hline
{\sc Task} & {\sc Approach} & {\sc 1 points} & {\sc 5 points} & {\sc 10 points} & {\sc \#Target (labels)}\\
\Xhline{0.5\arrayrulewidth}
\multirow{3}{*}{\shortstack{{\color{gg}Text}\\$\downarrow$\\{\color{rr}Image}}} & \multirow{1}{*}{Unimodal image} & $37.4 \pm 0.6$ & $41.7 \pm 3.7$ & $49.0 \pm 1.0$ & $5131 (0)$ \\
\cline{2-6}
& \multirow{1}{*}{\name\ (ours)} & $\mathbf{39.7 \pm 1.3}$ & $\mathbf{47.1 \pm 3.3}$ & $\mathbf{51.1 \pm 2.1}$ & $5131 (0)$ \\
\cline{2-6}
& \multirow{1}{*}{Oracle image~\cite{finn2017model,nichol2018reptile}} & $38.9 \pm 2.1$ & $42.1 \pm 1.4$ & $47.9 \pm 5.6$ & $5131 (5131)$ \\
\hline \hline
\end{tabular}

\vspace{2mm}

\begin{tabular}{C{1.7cm} L{3cm} | *{3}{C{1.6cm}} C{2.7cm}}
\hline \hline
\multirow{3}{*}{\shortstack{{\color{gg}Image}\\$\downarrow$\\{\color{rr}Audio}}} & \multirow{1}{*}{Unimodal audio} & $45.6 \pm 1.3$ & $74.2 \pm 0.3$ & $83.7 \pm 0.1$ & $920 (0)$ \\
\cline{2-6}
& \multirow{1}{*}{\name\ (ours)} & $\mathbf{47.5 \pm 0.2}$ & $\mathbf{85.9 \pm 0.7}$ & $\mathbf{92.7 \pm 0.4}$ & $920 (0)$ \\
\cline{2-6}
& \multirow{1}{*}{Oracle audio~\cite{finn2017model,nichol2018reptile}} & $45.9 \pm 0.2$ & $\mathbf{89.3 \pm 0.4}$ & $\mathbf{94.5 \pm 0.3}$ & $920 (920)$ \\
\hline \hline
\end{tabular}

\label{colearning}
\vspace{0mm}
\end{table*}

\textbf{Results:} From Table~\ref{colearning}, we observe that \name\ outperforms unimodal approaches. Unimodal approaches struggle due to only a small number of datapoints in the target modality.

Surprisingly, \name\ also manages to slightly outperform the oracle baseline on the text to image transfer task. We hypothesize this is because text data (source) is cleaner than image data (target) and these are the tasks where we have more total labeled data in the source modality (text) and less total labeled data in the target (image). Consistent with this hypothesis, we found that text classifiers performed better on the Yummly-28K dataset than image classifiers (in reference to Table~\ref{fusion}, where unimodal text gets $60.0\%$ while unimodal image gets $55.0\%$ accuracy). This implies that one can leverage abundant, cleaner, and more-predictive source modalities to improve target modality performance by learning a multimodal language. For image to audio (Table~\ref{colearning} middle), we observe that our approach is on par (outperforms for $k=1$, and within $2-3\%$ for $k=5,10$) with the oracle baseline that has seen a thousand labeled audio examples in the target modality.

\vspace{-1mm}
\subsection{Main Take-away Messages}
\vspace{-1mm}

From this case study of multimodal language learning in a computational model of artificial intelligence using real-world machine learning models and datasets~\citep{liang2020cross}, we find that the multimodal language we have implemented has successfully learned to perform several tasks such as multimodal fusion, alignment, and co-learning simultaneously. Performance is consistently superior to unimodal language learning on the fusion and co-learning tasks, while unimodal learning does not enable alignment at all. While we are unable to fully explore multimodal generation due to a lack of high-fidelity generators and evaluation metrics for image and audio, we leave this part for future work. Furthermore, it would also be interesting to integrate a similar multimodal language with an actual computational model of consciousness such as the Conscious Turing Machine (CTM)~\citep{blum2021theoretical}, and test it in simulated environments.

\vspace{-2mm}
\section{Conclusion}
\vspace{-2mm}

In conclusion, we formalized the properties of a multimodal language, \name, essential for machine models of intelligence and consciousness. We connected these properties to the core technical challenges and algorithms for multimodal representation learning from an AI perspective and proposed ideas towards operationally learning such a multimodal language. We hope that these insights can serve as a bridge between the study of multimodal representations in human and artificial intelligence with the eventual goal of developing a similar multimodal language needed to achieve intelligence and consciousness in artificial machines.

\textbf{Future directions:} We outline several important directions of future research:

1. Quantifying differences between modalities: One core challenge of multimodal learning lies in representing and synchronizing vastly heterogeneous modalities which require different unimodal processors. However, different sensory inputs are more similar than others. For example, speech and language could be seen as more similar than images and text. Furthermore, it is unclear if speech in different languages should be classified as being the same or different modalities. Future work should investigate formalisms of heterogeneity in multimodal data and how heterogeneity plays a role in the design decisions when learning a multimodal language.

2. Plasticity of unimodal processors in the brain: Unimodal processors are not static over time but rather evolve with our surroundings, especially when encountering lesions such as post-birth blindness~\cite{loiotile2019naturalistic}. Similarly, when acquiring new skills (e.g., learning a new language), new processors may develop in conjunction with existing ones~\cite{kuhl2011early}. Our investigation into a multimodal language has only explored static processors across a predefined number of input modalities and currently lacks the flexibility to handle dynamic modalities and processors.

3. Multimodal integration in the brain: While we have presented a multimodal language based on coordinated representations, insights from neuroscience regarding unimodal and multimodal processing, integration, alignment, translation, and co-learning~\cite{calvert2001crossmodal,kosslyn2010multimodal,nanay2018multimodal} could potentially inform our design of multimodal models.

4. Environments and evaluation: It is important to design simulated environments that are reflective of real-world human learning processes in order to benchmark design decisions in the multimodal language and AI models as a whole. These environments should capture the multimodality of diverse environments, a comprehensive suite of possible agent interactions, and feedback signals at varying granularities, all while remaining efficient and reproducible.

\vspace{-2mm}
\section*{Acknowledgements}
\vspace{-2mm}

PPL is supported in part by a Facebook PhD Fellowship and a Carnegie Mellon University's Center for Machine Learning and Health Fellowship. Any opinions, findings, conclusions, or recommendations expressed in this material are those of the author(s) and do not necessarily reflect the views of Facebook or Carnegie Mellon University's Center for Machine Learning and Health, and no official endorsement should be inferred. The authors are extremely grateful to Manuel Blum and Lenore Blum for encouraging the development of this manuscript as well as helpful discussions and feedback on computational models for artificial intelligence and consciousness. We also thank Louis-Philippe Morency and Ruslan Salakhutdinov, as well as the students in their research groups (Alex Wilf, Amir Zadeh, Ben Eysenbach, Brandon Trabucco, Chaitanya Ahuja, Dong Won Lee, Martin Ma, Murtaza Dalal, Peter Wu, Tiffany Min, Torsten Wortwein, Victoria Lin, Volkan Cirik, Yao-Hung Hubert Tsai, Yiwei Lyu) for constructive discussions regarding multimodal machine learning. We acknowledge Peter Wu for his role in the implementation and experiments in Section~\ref{sec:ml}. Finally, we would also like to acknowledge NVIDIA's GPU support.

\clearpage

{\small
\bibliography{refs}
\bibliographystyle{plainnat}
}

\end{document}